%% file: main.tex
\documentclass{article}

\usepackage[final]{neurips_2025}

\usepackage[utf8]{inputenc} %
\usepackage[T1]{fontenc}    %
\usepackage{hyperref}       %
\usepackage{url}            %
\usepackage{booktabs}       %
\usepackage{amsfonts}       %
\usepackage{nicefrac}       %
\usepackage{microtype}      %
\usepackage{xcolor}         %
\hypersetup{colorlinks=true,linkcolor=blue,citecolor=blue}

\title{Improving planning and MBRL with temporally-extended actions}

\usepackage{amssymb}            %
\usepackage{mathtools}          %
\usepackage{mathrsfs}           %
\usepackage{graphicx}           %
\usepackage{subcaption}         %
\usepackage[space]{grffile}     %
\usepackage{url}                %
\usepackage{lipsum}             %

\input{custom_commands}
\definecolor{ablue}{RGB}{31,119,180}
\definecolor{aorange}{RGB}{255,127,3}

\usepackage{amsmath}
\usepackage{amsfonts}
\usepackage{cleveref}
\usepackage{algorithm}
\usepackage{algpseudocode}
\usepackage{nicefrac}
\usepackage{booktabs}
\usepackage{multirow}
\usepackage{dsfont}
\usepackage{bm}
\usepackage{tikz}
\usepackage{siunitx}
\usepackage{titletoc}
\usepackage{wrapfig}
\usepackage{dsfont}
\usepackage[inline]{enumitem}
\usepackage{todonotes}

\author{%
  Palash Chatterjee \\
  Indiana University \\
  \texttt{palchatt@iu.edu} \\
  \And Roni Khardon \\
  Indiana University \\
  \texttt{rkhardon@iu.edu} \\
}

\begin{document}

\maketitle

\begin{abstract}
Continuous time systems are often modeled using discrete time dynamics but this requires a small simulation step to maintain accuracy.
In turn, this requires a large planning horizon which leads to computationally demanding planning problems and reduced performance. 
Previous work in model-free reinforcement learning has partially addressed this issue using action repeats where a policy is learned to determine a discrete action duration.
Instead we propose to control the continuous decision timescale directly by using \TEAs~ and letting the planner treat the duration of the action as an additional optimization variable along with the standard action variables. 
This additional structure has multiple advantages. 
It speeds up simulation time of trajectories and, importantly, it allows for deep horizon search in terms of primitive actions while using a shallow search depth in the planner. 
In addition, in the model-based reinforcement learning (MBRL) setting, it reduces compounding errors from model learning and improves training time for models. 
We show that this idea is effective and that the range for action durations 
can be automatically selected using a multi-armed bandit formulation and integrated into the MBRL framework.
An extensive experimental evaluation both in planning and in MBRL, shows that our
approach yields faster planning, better solutions, and that it enables solutions to problems that are not solved in the standard formulation.
\end{abstract}

\input{sections/introduction/main}
\input{sections/background/main}

\input{sections/method/main}

\input{sections/experiments/main}

\input{sections/conclusion/main}

\input{sections/future_work}
\input{sections/ack.tex}

\bibliographystyle{plainnat}
\bibliography{ref}

\newpage
\input{sections/checklist}

\newpage
\appendix
\input{appendix/main}

\end{document}

%% file: custom_commands.tex
\newcommand{\Smdp}{\mathcal{S}}
\newcommand{\A}{\mathcal{A}}
\newcommand{\R}{\mathcal{R}}

\newcommand{\Ex}{\mathbb{E}}
\newcommand{\T}{\mathcal{T}}

\newcommand{\timescale}{\delta_t}
\newcommand{\dtmin}{\delta t_{\min}}
\newcommand{\dtmax}{\delta t_{\max}}
\newcommand{\dtenv}{\delta_t^{\text{env}}}
\newcommand{\dtk}{\delta t_k}
\newcommand{\dt}{\delta t}

\newcommand{\tea}{\text{TE}}
\newcommand{\naive}{\text{STD}}

\newcommand{\var}{\text{var}}
\newcommand{\ip}{\text{IP}}
\newcommand{\fip}{F_{\text{IP}}}
\newcommand{\fhatip}{\hat{F}_{\text{IP}}}
\newcommand{\fhattea}{\hat{F}_{\tea}}

\newcommand{\TE}{temporally-extended}
\newcommand{\TEA}{temporally-extended action}
\newcommand{\TEAs}{temporally-extended actions}
\newcommand{\TEAspace}{temporally-extended actions }

%% file: sections/introduction/main.tex
\section{Introduction}

Many interesting real life systems evolve continuously with time, but the dynamics of such systems are often modeled using a discrete-time approximation. 
In these models, time evolves in discrete steps of $\timescale$ called the \emph{timescale} of the system. 
Simulators used in RL or robotics often use such models to capture physical systems.
Setting $\timescale$ to a small value allows the discrete-time models to have a good local approximation.
To obtain a trajectory of the system using such models, the dynamics function needs to be evaluated at fixed intervals of $\timescale$.
Because $\timescale$ is small, the number of decisions required to solve even a simple task can be quite large. 
From the perspective of planning or MBRL, this translates to longer planning horizons (or rollouts) which can limit their effectiveness.

Shooting based planners like Cross Entropy Method (CEM) and Model Predictive Path Integral (MPPI) \citep{kobilarov2012cross,botev2013cross,williams2017model,chua2018deep} rely on sampling to estimate the highly rewarding regions of the state space. 
They are known to perform poorly as the planning horizon increases, especially in environments with noisy dynamics or sparse rewards. This is because in such environments the variance of their estimates increases and they need a large number of samples in order to act reliably \citep{chatterjee2023disprod}.  In addition, when planning with learned models as in MBRL, 
longer rollouts can lead to compounding errors \citep{janner2019mbpo}.

Simulators like MuJoCo \citep{todorov2012mujoco} or Arcade Learning Environment \citep{bellemare13arcade} make use of a fixed \emph{frame-skip} in addition to timescale. A frame-skip or \emph{action repeat} of $n$ means that the same action is repeated for $n$ times before the agent is allowed to act again. So the \emph{decision timescale} becomes $n \times \timescale$. 
Frame-skip values are normally set heuristically, but as shown by \citet{braylan2015frame}, the ideal values of frame-skip vary depending on the environment and can heavily influence the performance of the agent.
This has led to many efforts in trying to control the decision timescale by learning a policy to control either frame-skip \citep{durugkar2016deep,lakshminarayanan2017dynamic,sharma2017learning},  or timescale \citep{ni2022continuous}. 
To the best of our knowledge, none of the previous approaches modeling the decision timescale consider the planning problem or the use of planning in MBRL.

In this paper, we propose to control the decision timescale directly by treating the action duration $(\dt)$ as an additional optimization variable.
At each \emph{decision step},  the planner optimizes for the standard \emph{primitive action} as well as its duration
resulting in a \emph{\TEA}.  
In contrast to using a fixed frame-skip, the duration for each \TEA~ can be different allowing the planner to choose actions of varying durations, as can be seen in \Cref{fig:histogram_of_action_repeats}.
This provides the planner with an added degree of flexibility and, at the same time, it reduces the search space by constraining the structure of the trajectories that the planner searches over.
Further, as a small increase in the \TE~ planning horizon can translate to a large increase in the primitive planning horizon, this allows the agent to search deeper, enabling it to solve environments which are otherwise too difficult.

Finally, unlike previous work, we do not learn a policy. 
Our work is in the context of MBRL. We learn a model that predicts the next state and reward due to a \TEA~ and use the learned model to plan, leading to superior performance as well as faster training. 
Selecting a suitable range for $\dt$ is an important step as it can directly impact the performance of the planner.
We show that posing this as a \emph{non-stationary} multi-armed bandit (MAB) problem \citep{besbes2014stochastic,lattimore2020bandit} allows us to select the range automatically.

To summarize, our main contributions are as follows:
\begin{enumerate}[leftmargin=*]
    \item We propose to control the decision timescale directly by letting the planner optimize for primitive action variables as well as the duration of the action, and evaluate this idea both in planning and in MBRL. 

    \item We show that our approach decreases the search space and the planning horizon for the planner, helping it solve previously unsolvable problems. 
    It also improves the stability of the search by allowing the planner to search deeper and solve problems with sparse rewards.
   
    \item 
    We show that learning \TE~ dynamics models and selecting a suitable range for $\dt$ can be integrated into a single algorithm within the MBRL framework, and that planning using these learned models leads to faster training and superior performance over the standard algorithm.
\end{enumerate}

%% file: sections/background/main.tex
\input{sections/background/related_work}

%% file: sections/background/related_work.tex
\section{Related Work}

\emph{Macro-actions} \citep{finkelstein1998selective,hauskrecht2013hierarchical} and  \emph{options} \citep{sutton1999between} are two of the most common methods to introduce temporal abstraction in planning and RL. 
A macro-action is a usually defined to be sequence of primitive actions that the agent will take. On the other hand, an option consists of a policy, an initiation set and a terminating condition. The initiation set determines when the option can be taken, while its duration depends on when its terminating condition is satisfied. Both macro-actions and options can either be predefined or can be learned from data \citep{durugkar2016deep,machado2017laplacian,machado2017eigenoption,ramesh2019successor}.

Frame-skips or action-repeats are simpler forms of macro-actions where each macro-action is essentially a single primitive action repeated multiple times. While frame-skipping has been used as a heuristic in many deep RL solutions \citep{bellemare2012investigating,mnih2015human}, \citet{braylan2015frame} showed that using a static value of frame-skips across environments can lead to sub-optimal performance. 
Following this, there have been multiple efforts using the model-free RL framework to make the frame-skip dynamic.
\citet{lakshminarayanan2017dynamic} propose to learn a joint policy on an inflated action space of size $n|\A|$ where each primitive action is tied to $n$ corresponding action-repeats. The value of $n$ is usually small to limit the size of the inflated action space. 
A more practical approach is to learn a separate network alongside the standard policy to predict the number of action repeats \citep{sharma2017learning,biedenkapp2021temporl}.

Frame-skips or action-repeats are discrete values and, while they introduce a temporal abstraction for discrete time systems, 
a continuous representation is both more realistic and more flexible. 
Some prior work in model-free RL has explored this problem. 
 \citet{ni2022continuous} use a modified Soft Actor Critic \citep{haarnoja2018soft} to learn a policy and control the timescale rather than the action duration, with an explicit constraint on the average timescale of the policy. 
 Their formulation of the optimization objective is similar to ours except that they use a linear approximation of the reward function to compute the reward due to a macro-action (details in \Cref{sec:connection_to_continous_control_on_time}).
 \citet{wang2024deployable} introduce Soft-Elastic Actor Critic to output the duration of the action along with the action itself. However, they modify the reward structure by introducing penalties for the energy and the time taken by each action. 
 Our work is in the planning and MBRL setting. 
 Rather than learn a policy, we use either exact or learned transition and reward functions to plan. Our objective arises naturally from the formulation without the need to add additional constraints or engineer rewards. 
 Further, previous work in model-free RL treat the maximum value of action-repeats or timescale as a hyperparameter.
 Instead, we use a MAB framework, similar to \cite{li2018hyperband} and \cite{lu2022non}, and automatically select the maximum action duration removing the need for tuning an additional hyperparameter. 

Finally, in our work, action durations are chosen by the planner, which is different than planning with durative actions \citep{MausamW08} where the durations are given by the environment.

%% file: sections/method/main.tex
\section{Modeling Temporally-Extended Actions}
\input{sections/background/background}

Although we eventually care about what \emph{primitive actions} to take in the environment, a planner can work at an abstract level by using \emph{\TEAs}. We use the terms \emph{decision steps} and \emph{execution steps} to distinguish between the number of times the agent outputs an action and the number of \emph{primitive actions} that are actually executed in the environment. 

Let us assume that the agent has access to the \emph{primitive dynamics function} $(f)$ which is accurate for all $0\le t \le \dtenv$. 
In the standard setup, the agent uses $f$ at each decision step and outputs an action whose duration is implicitly $\dtenv$. The return due to a trajectory $\tau$ is given by $J_{1} = \sum_{t=1}^{L(\tau)} \gamma^{t-1} \R(s_t, a_t)$
Here, the number of execution steps is exactly equal to the number of decision steps.

In our proposed framework, the planner explicitly outputs the duration of the action along with the action itself. At decision step $k$, let the planner output an action $a_k$ and its corresponding duration $\dtk \in [\dtmin, \dtmax]$. Now the number of execution steps need not necessarily be equal to the number of decision steps. Let $e_{k} = \lfloor{\nicefrac{\dtk}{\dtenv}}\rfloor + \mathds{1}(\dtk~\texttt{mod}~\dtenv)$ be the number of execution steps associated with decision step $k$. Further, let $e_{<k} = \sum_{j=1}^{k-1} e_j$ be the total number of execution steps taken prior to decision step $k$. The return due to a trajectory $\tau$ will be 
\begin{align}
    \label{eq:proposed_objective_j2}
    J_{2} = \sum_{k = 1}^{L(\tau)} \gamma^{e_{<k}} \sum_{t = 1}^{e_k} \gamma^{t-1} \R(s_{(e_{<k} + t)}, a_{(e_{<k} + t)}) 
\end{align}
where $e_{<1} = 0$. This formulation provides an approximation that avoids the need for a precise continuous time model of discounting. Note that when $\dtk = \dtenv$ for all $k$, then $J_{2} = J_{1}$.

Let $R^{\tea}_{k} = \sum_{t = 1}^{e_k} \gamma_2^{t-1} \R(s_{(e_{<k} + t)}, a_{(e_{<k} + t)})$ be the reward due to the temporally-extended action at decision-step $k$. Then we have a slightly different view of $J_2$ where the returns due to a temporally-extended action is discounted based on the number of primitive actions taken prior to the current timestep. Formally,
\begin{align}
    \label{eq:proposed_objective}
    J_{3} &= \sum_{k = 1}^{L(\tau)} \gamma_1^{e_{<k}} R^{\tea}_{k} = \sum_{k = 1}^{L(\tau)} \gamma_1^{e_{<k}} \biggl( \sum_{t = 1}^{e_k} \gamma_2^{t-1} \R(s_{(e_{<k} + t)}, a_{(e_{<k} + t)}) \biggr)
\end{align}
This view allows us to have different discount factors, $\gamma_1$ and $\gamma_2$, giving us more fine-grained control over the behavior of the agent.

\input{algorithm/planning_use_fte}

\section{Method}

We want the planner to have access to a \emph{\TE {} dynamics function} $(F)$ that can work with \TEAs. If $F$ is available, then using it for planning is straightforward.
For example, one can use a shooting-based planner with $F$ to select an action from any state $s$ as shown in  \Cref{alg:using_fte_with_shooting_planner}. However, $F$ is usually not readily available. In the standard planning setup, the agent has access to the primitive dynamics function $(f)$ which samples from the one step primitive transition and reward distribution.
A simple solution is to wrap $f$ in a loop (as shown in \Cref{alg:long_horizon_transition_function} in the Appendix) to obtain $\fip$. This holds as we assumed that $f$ is accurate for all $0\le t \le \dtenv$.
We call this the \emph{iterative primitive dynamics function}.
Although $\fip$ accurately captures $F$, it fails to facilitate the speed of execution.
The time required by $\fip$ to simulate the outcome due to a \TEA~ is dependent on the duration of the action itself. However, if we had access to $F$, this would have been a constant time evaluation.

\textbf{Approximating $F$ :} 
We address this by using neural networks to approximate $F$ using $\fhattea$. 
This allows us to predict the next state and reward due to a temporally-extended action from a given state in constant time.
For learning this model, we use the framework of MBRL. We collect data by interacting with the environment and use the data to train $\fhattea$ 
to predict a distribution over the next states and a point estimate for the reward. 
The learned model can then be used for planning. For example, \Cref{alg:using_fte_with_shooting_planner} can be used with $\fhattea$ instead of $F$ to select actions, as we do in this work.

\textbf{Selecting $\dtmin$ and $\dtmax$ :}
Unlike other action variables, the range of $\delta t$ does not come from the environment and has to be set by the planner. 
If the range is too large, the search for the optimal $\delta t$ becomes harder while a smaller range can limit the search. 
However, as the range simply influences the spread of the distribution used to sample $\dt$, its value need not be set precisely.

We set $\dtmin = \dtenv$ to be the timescale of the environment. 
For $\dtmax$, we consider $m$ exponentially-spaced candidates, where $m = \log_2(T)$ and $T$ is the maximum number of primitive steps defined by the environment. We model each $\dtmax$ candidate as an arm and pose the selection as a MAB problem.
Pulling an arm is equivalent to selecting one of the $\dtmax$ candidates and using the planner with this value of $\dtmax$ to collect data for one episode. The aggregate reward observed at the end of an episode indicates the quality of the arm. 
However, note that the rewards obtained are directly related to the quality of the learned dynamics model the agent has. As the agent collects more data, the model quality and hence the reward distribution of the arms change, making the problem of arm selection non-stationary.
To address this we propose to use an exponential moving average that focuses the reward estimate on recent episodes and hence ameliorates the effect of non-stationarity.

Let $\bar{R}_{T}$ be the average reward per timestep obtained in episode $T$ and let $\hat{R}_{i,T} = \hat{R}_{i,T-1} + \alpha (\bar{R}_{T} - \hat{R}_{i,T-1} )$ be the exponentially moving average (EMA) of rewards per timestep obtained till episode $T$ due to candidate $i$. 
Let $N(i,T)$ be the number of times candidate $i$ was chosen before iteration $T$.  
At the start of iteration $T+1$, $\dtmax$ is selected using the UCB heuristic \citep{auer2002finite} in \Cref{eq:ucb}
where we set $\hat{R}_{i,T=0}$ to the mean reward obtained across 5 episodes using the randomly initialized dynamics models, prior to any training.
\begin{align}
\label{eq:ucb}
    \arg \max_i \biggl(\hat{R}_{i,T} + c \sqrt{\frac{2 \log(T)}{N(i,T)}}\biggr)
\end{align}

Another important question is whether to learn a single dynamics model to be shared by all the arms or learn a separate dynamics model for each arm. Our preliminary experiments, which we describe in \Cref{sec:appendix_single_model_vs_sep_model}, suggested that learning a separate dynamics model for each $\dtmax$ candidate is better.

To summarize, we treat each $\dtmax$ candidate as a bandit arm. Before each episode, we use a UCB heuristic with EMA of rewards to select an arm and use the selected value of $\dtmax$ and the corresponding dynamics model to plan for the duration of the episode.

\textbf{Discussion :}
Consider two agents - $A_{\naive}$ that uses primitive actions and $A_{\tea}$ that uses temporally-extended actions and let  $D_{\naive}$ and $D_{\tea}$ be their corresponding planning horizons. Further, for $A_{\tea}$, let $\dtmax = m \times \dtenv$ for some positive integer $m$.

The planning horizon for the two agents are at different scales and not directly comparable. We introduce the term \emph{maximal primitive horizon} $(H)$ which is the maximal number of primitive actions taken by the agents while planning. For $A_{\naive}$, $H = D_{\naive}$, while for $A_{\tea}$, $H = m \times D_{\tea}$. 
A fixed value of $H$ ensures that, while planning, $A_{\tea}$ does not consider more primitive actions than $A_{\naive}$, at any instant. For practical purposes, $A{_\tea}$ will take less primitive actions than $A_{\naive}$. We argue that even in this unfair setting, using temporally-extended actions can be beneficial.
\begin{enumerate}[leftmargin=*]
    \item 
    Using temporally-extended actions helps to reduce the space of possible trajectories that $A_{\tea}$ searches over. For simplicity, let us consider an environment with binary actions. 
    The search space for $A_{\naive}$ is $2^H$, while for $A_{\tea}$, the search space reduces to $2^{H/m}$. 
    However, this reduction comes at a cost of the flexibility of trajectories
    as when $m$ is larger the generated trajectories have a more rigid structure.
    \item 
    $A_{\tea}$ needs to optimize less variables than $A_{\naive}$. In general, if the action space has dimensions $|\A|$, then at each decision step, $A_{\naive}$ has to optimize for $H |\A|$ action variables, while $A_{\tea}$ will have to optimize for $(H/m)(|\A|+1)$ action variables. 
    \item
    $\fhattea$ evaluates the outcome of temporally-extended actions in constant time, which is similar to that taken by $f$ to evaluate the outcome of a primitive action. As $D_{\tea} < D_{\naive}$ for all practical purposes, using $\fhattea$ with $A_{\tea}$ helps it make a decision faster than $A_{\naive}$. 
    \item 
    Manipulating the value of $m$ allows us to scale up the maximal planning horizon of $A_{\tea}$ without adding any extra variables. This can be useful in environments where 
    rewards are uninformative and 
    a deeper search is required.
\end{enumerate}

%% file: sections/background/background.tex
The standard discrete-time Markov Decision Process (MDP) is specified by $\{\Smdp, \A, \T, \R, \gamma \}$ where $\Smdp$ and $\A$ are the state and action spaces respectively and  $\gamma \in (0, 1)$ is the discount factor. $s_t \in \Smdp$ and $a_t \in \A$ represents the state and \emph{primitive action} at timestep $t$. 
The one-step transition distribution is given by $\T(s_{t+1}| s_t, a_t)$ and the one-step reward distribution is given by $\R(s_t,a_t)$. 
The expected discounted return is given by $J_t = \Ex \biggl[ \sum_{i=0}^{D-1} \gamma^{i} \R(s_{t+i}, a_{t+i}) \biggr]$ where $D$ is the planning horizon.
Discrete time simulation of continuous systems assumes that $\T$ and $\R$ capture the transitions and the corresponding reward due to exactly $\dtenv$ duration, where $\dtenv$ is the timescale of the MDP.
In cases when $\T$ and $\R$ are unknown to the agent, their empirical estimates, $\hat{\T}$ and $\hat{\R}$, can be learned using data collected by interacting with the MDP.
For the following discussion, we overload the term dynamics to mean both the transition and reward function.

%% file: algorithm/planning_use_fte.tex
\begin{algorithm}[tbp]
\caption{One decision step : Action selection using a \TE {} dynamics function ($F$) with a generic shooting-based planner}\label{alg:using_fte_with_shooting_planner}
\begin{algorithmic}[1]
\Require \TE~ dynamics function $(F)$, current state $(s)$ \newline
\hspace*{2.2em} initial action distribution $(\mu_a, \var_a)$, number of rollouts $(N)$, planning horizon $(D_{\tea})$
\For{\text{$i = 1$ to optimization steps}}
\State Sample $N$ action sequences of length $D_{\tea}$ using $\mu_a, \var_a$
\For{\text{all $N$ action sequences $\textbf{a}_{t:t+D_{\tea}} = (a_t, \dots, a_{t + D_{\tea}})$}}
\State \textcolor{blue}{\texttt{//$a_t$ includes standard action variables and the action duration}}
\State \text{Simulate trajectory using $F$ from $s$ due to $\textbf{a}_{t:t+D_{\tea}}$ and compute aggregate reward}
\EndFor
\State \textcolor{blue}{\texttt{//update rule is specific to the planner being used.}}
\State Update $\mu_a, \var_a$ using action sequences and aggregate rewards
\EndFor  
\State \Return sample action using $\mu_a, \var_a$
\end{algorithmic}
\end{algorithm}

%% file: sections/experiments/main.tex
\section{Experiments}

We evaluate the use of temporally-extended actions in planning and in MBRL. We begin by elaborating on our method for both of these settings, before discussing the experiments and results. 

\textbf{Experimental setup : } We compare the planning performance of $A_{\naive}$ and $A_{\tea}$ using CEM, which is a shooting-based planner.
CEM maintains a sequence of sampling distributions from which it generates multiple action sequences. For each action sequence, it instantiates multiple particles and computes the trajectory and reward due to each particle. Then, it computes the mean reward per action sequence and uses the top $k$ action sequences to bias its sampling distributions. 
For evaluation, we wrap the primitive dynamics function of the simulation environment in an iterative loop using $\fip$ as discussed in the previous section. 

For experiments with RL, following \citet{chua2018deep},
we learn an ensemble of neural networks to approximate the dynamics function. Specifically, $A_{\naive}$ learns to approximate $f$ while $A_{\tea}$ learns to approximate $F$. Both the agents use the same network architecture and configuration of parameters. 
The only difference is that the model learned by $A_{\tea}$ has an extra input variable corresponding to $\delta t$, in addition to the state and action variables. 
The data collection and the training procedure for $A_{\tea}$ with a fixed $\dtmax$ in described in \Cref{alg:training_loop}. We use the TS$\infty$ variant from \cite{chua2018deep} for planning with the learned ensemble of dynamics models (see \Cref{sec:hyperparameters} for details about planning with ensembles). Note that $A_{\naive}$ is equivalent to TS$\infty$ from PETS \citep{chua2018deep}.
\input{algorithm/training_loop.tex}

We consider two variants of our proposed algorithm - $A_{\tea}(\text{F})$ which uses a fixed and manually chosen value of $\dtmax$ and $A_{\tea}(\text{D})$ which uses the proposed MAB framework to dynamically select the value of $\dtmax$ at the beginning of each episode. 
Both variants use $\dtmin = \dtenv$.
A key difference between the two is that $A_{\tea}(\text{D})$ maintains a separate dataset and learns a separate temporally-extended dynamics model for each $\dtmax$ candidate. 
Specifically, $A_{\tea}(\text{F})$ uses \Cref{alg:training_loop} exactly, while $A_{\tea}(\text{D})$ has an additional step where it uses the MAB framework to select a $\dtmax$ value before the start of each episode and then uses the corresponding dataset and model for the duration of the episode.
For our discussion, we use $A_{\tea}$ to refer to both the variants in general while specifying the variant wherever required.

\subsubsection*{Experimental Evaluation} 
First, we experiment in a planning regime where the agent has access to the exact dynamics function. 
In this setting, $A_{\tea}$ uses iterative primitive dynamics function (\Cref{alg:long_horizon_transition_function}) for planning. 
For experiments, we use the Mountain Car environment from Gymnasium \citep{kwiatkowski2024gymnasium}, a multi-hill Mountain Car environment from the Probabilistic and Reinforcement Learning Track of the International Planning Competition (IPC) 2023 \citep{ipc2023}, and the Dubins car environment from \cite{chatterjee2023disprod}. 
Then, we experiment in the MBRL setting where $\T$ and $\R$ are not known. In this case, we learn a temporally-extended model as discussed above, by interacting with the environment. We use Cartpole from Gymnasium and Ant, Half Cheetah, Hopper, Reacher, Pusher and Walker from MuJoCo \citep{todorov2012mujoco}. We run each experiment across 5 different seeds and aggregate the results. Full details of the experimental setup are given in \Cref{sec:hyperparameters}.

The experiments are organized so as to answer a set of questions as outlined below.

\input{sections/experiments/exp_perf_vs_depth}

\input{sections/experiments/exp_dubins_u_maps}

\input{sections/experiments/exp_varying_agents_behaviour}

\input{sections/experiments/exp_online_learning}

%% file: algorithm/training_loop.tex
\begin{algorithm}[tbp]
\caption{Using $A_{\tea}$ with a fixed $\dtmax$ as planner for MBRL}\label{alg:training_loop}
\begin{algorithmic}[1]
\Require  dynamics model ensemble$(\hat{F}_{\text{TE}})$, batch size $(b)$, number of iterations $(n)$, $\dtmax$
\State Initialize empty dataset $\mathcal{D}$
\For{\text{$i = 1$ to $n$}}
\State \textcolor{blue}{\texttt{//data collection}}
\For{\text{$t = 1$ to end of episode}}
\State \textcolor{blue}{\texttt{//$a_t$ includes standard action variables and the action duration}}
\State $a_t$ = \texttt{planner.choose\_action($s_t$, $\dtmax$)}
\State \textcolor{blue}{\texttt{//env.step() implicitly uses temporally-extended dynamics}}
\State $s_{t+1}, r_t$ = \texttt{env.step($a_t$)}
\State \texttt{$\mathcal{D}$ = $\mathcal{D}$.append(($s_t, a_t, s_{t+1}, r_t$))}
\EndFor
\State \textcolor{blue}{\texttt{//training}}
\State \texttt{num\_of\_batches = len($\mathcal{D}$) / $b$}
\For{\text{$j = 1$ to \texttt{num\_of\_batches}}}
\For{\text{each dynamics model in $\fhattea$}}
    \State Sample a mini-batch of size $b$ from $\mathcal{D}$ \label{alg_line:sample_mini_batch}
    \State Compute the loss and gradients using the mini-batch
    \State Update the model using the gradients
\EndFor
\EndFor
\EndFor  
\end{algorithmic}
\end{algorithm}

%% file: sections/experiments/exp_perf_vs_depth.tex
\textbf{Does planning with temporally-extended actions help?}
Mountain Car has a sparse reward, requiring a large planning horizon to succeed. 
We compare the performance of $A_{\naive}$ and $A_{\tea}$ by varying the planning horizon. As shown in \Cref{fig:exp_perf_vs_depth}, $A_{\tea}(D_{\tea} \ge 4)$ solves the environment with a much smaller planning horizon than $A_{\naive}(D_{\naive} \ge 60)$.

\input{images/exp_perf_vs_depth/main}

Next, we experiment with the multi-hill version of Mountain Car from IPC 2023 (illustrated in \Cref{fig:ipc_mountain_car_example} in the Appendix). Each instance of the environment increases the difficulty by either adding more hills or altering the surface of the hills. 
As shown in \Cref{tab:exp_icaps_mc}, 
while $A_{\naive}$ is able to solve only the first instance, $A_{\tea}$ solves all the instances. 
Note that, even though $A_{\tea}$ takes a small number of decision steps, the computation time required to finish an episode is comparable to $A_{\naive}$. 
This is because $A_{\tea}$ uses an iterative version of the dynamics $(F_{\ip})$ for simulation.
Another observation is that because the episode terminates when the agent reaches the goal, the number of decision steps is smaller than $D_{\tea}$.

\input{tables/exp_perf_vs_depth/icaps_mc_undiscounted_15_vs_150}

%% file: images/exp_perf_vs_depth/main.tex
\begin{wrapfigure}{R}{0.5\textwidth}
    \centering%
     \subfloat{%
        \includegraphics[width=0.8\linewidth]{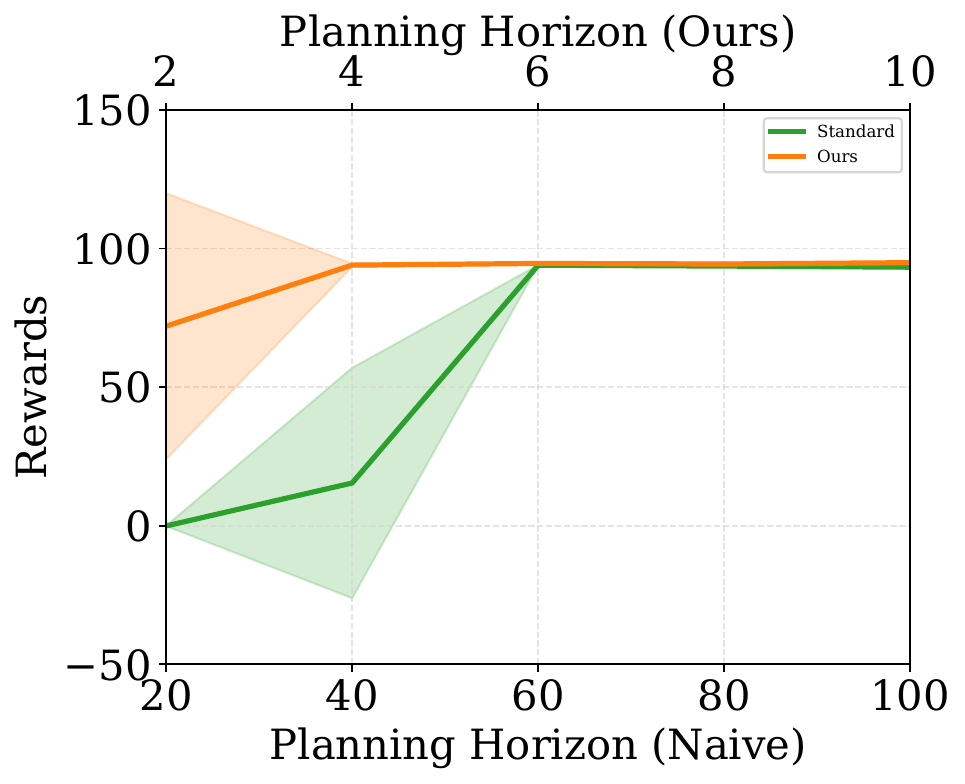}%
    }
    \caption{$A_{\naive}$ requires a large planning horizon $(D_{\naive} \ge 60)$ to succeed in Mountain Car, but $A_{\tea}$ using $\dtmax=100$ can work with a small planning horizon $(D_{\tea} \ge 4)$.}%
    \label{fig:exp_perf_vs_depth}
\end{wrapfigure}

%% file: tables/exp_perf_vs_depth/icaps_mc_undiscounted_15_vs_150.tex
\begin{table*}[t]
    \centering
    \scriptsize
 \begin{tabular}{lcccccccc}
\toprule
        Instance & \multicolumn{2}{c}{Rewards $\uparrow$ } & \multicolumn{2}{c}{Decision steps $\downarrow$ } & \multicolumn{2}{c}{Time for an episode $\downarrow$ } & \multicolumn{2}{c}{Success Probability $\uparrow$ } \\
 \cmidrule{2-9}
                    & Standard &  Ours                  & Standard &  Ours                    & Standard & Ours  &  Standard & Ours\\
\midrule
1 & 90.98 $\pm$ 0.26    & $91.74 \pm 1.55$  & 108.8 $\pm$ 0.84 & 3.0 $\pm$ 1.41 & 4.86 $\pm$ 0.19 & 5.7 $\pm$ 1.13  & 1.00 & 1.00 \\
2 &  -0.1 $\pm$ 0.01    & $\bm{88.67 \pm 0.65}$  & 300.0 $\pm$ 0.0  & 2.8 $\pm$ 1.1  & 7.98 $\pm$ 0.28 & 5.59 $\pm$ 0.91 & 0.00 & $\bm{1.00}$ \\
3 &  -0.1 $\pm$ 0.01    & $\bm{86.15 \pm 1.11}$  & 300.0 $\pm$ 0.0  & 2.6 $\pm$ 0.89 & 7.94 $\pm$ 0.15 & 5.44 $\pm$ 0.76 & 0.00 & $\bm{1.00}$ \\
4 &  -0.1 $\pm$ 0.01    & $\bm{66.50 \pm 43.54}$  & 300.0 $\pm$ 0.0  & 3.2 $\pm$ 1.3  & 7.73 $\pm$ 0.28 & 6.01 $\pm$ 1.05 & 0.00 & $\bm{0.80}$\\
5 &  -0.1 $\pm$ 0.01    & $\bm{83.41 \pm 0.54}$  & 300.0 $\pm$ 0.0  & 3.0 $\pm$ 0.0  & 8.0  $\pm$ 0.28  & 5.82 $\pm$ 0.2  & 0.00 & $\bm{1.00}$ \\
\bottomrule
\end{tabular}
\caption{Results on Multi-hill Mountain Car from IPC 23 across 5 seeds where $A_{\tea}$ $(D_{\tea}=12)$ solve all instances while $A_{\naive}$ $(D _{\naive}=175)$ can only solve 1/5.} 
\label{tab:exp_icaps_mc}
\end{table*}

%% file: sections/experiments/exp_dubins_u_maps.tex
\input{images/exp_u_maps/main}

\textbf{Can temporally-extended actions help transform infeasible problems to feasible ones?}
We use the Dubins Car environment $(\dtenv = 0.2)$ and experiment with u-shaped map as shown in \Cref{fig:exp_u_maps}. 
This is a challenging configuration, where shooting-based planners often fail, because the car is initially facing the obstacles and a naive forward search hits the obstacles and does not yield useful information.
In addition,
as the reward is sparse, this map requires the planning horizon to be large. 
To solve the environment, $A_{\naive}$ requires 10,000 samples with a planning horizon of 1000, while $A_{\tea}$ requires a planning horizon of 75 and $\dtmax = 20$ (see detailed discussion in \Cref{sec:appendix_exp_u_maps}).

To increase difficulty, we augment the action space 
with 100 dummy action variables. Although these variables do not contribute to the dynamics or rewards, the agent is unaware of this and has to account for all the action variables. 
We show that in this difficult setting as well,
$A_{\tea}$ succeeds while $A_{\naive}$ fails due to large memory requirements. A simple computation shows that $A_{\naive}$ needs around 4GB of memory to 
track the sampled actions, whereas $A_{\tea}$ requires just 103MB. 
In other configurations that reduce the memory requirements (detailed in \Cref{sec:appendix_exp_u_maps}), the search of $A_{\naive}$ fails to find the goal.
This shows that using temporally-extended actions can often turn infeasible search problems into feasible ones.

%% file: images/exp_u_maps/main.tex
\begin{figure*}[t] 
    \centering%
    \begin{subfigure}[b]{0.25\textwidth}
         \centering
           \includegraphics[width=\linewidth]{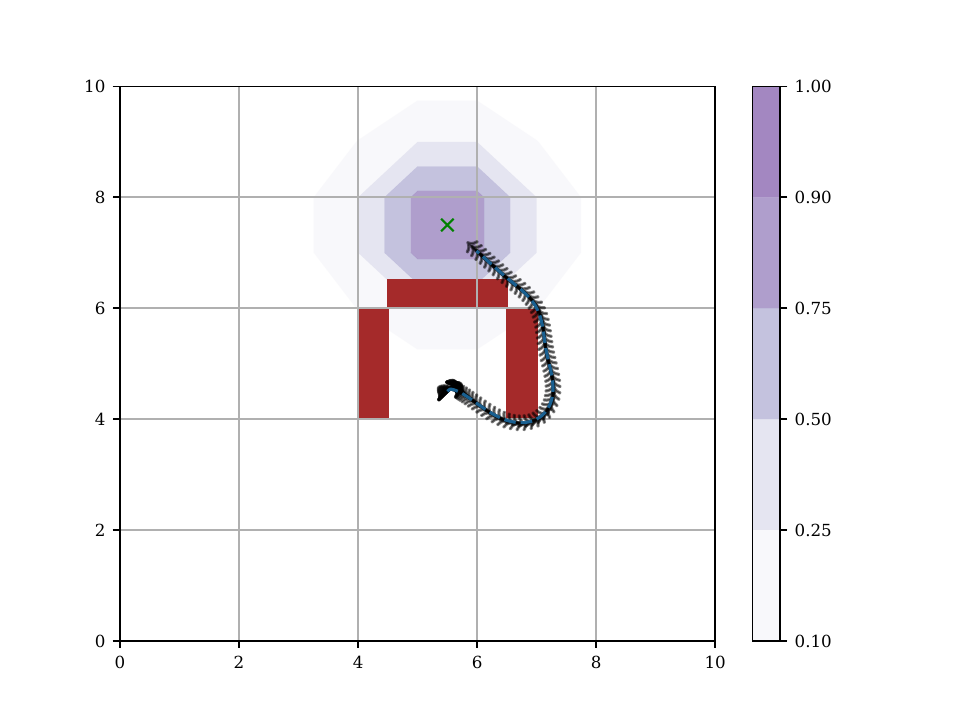}%
     \caption{$A_{\naive}, |\A|{=}2$}
    \end{subfigure}
    \begin{subfigure}[b]{0.25\textwidth}
         \centering
           \includegraphics[width=\linewidth]{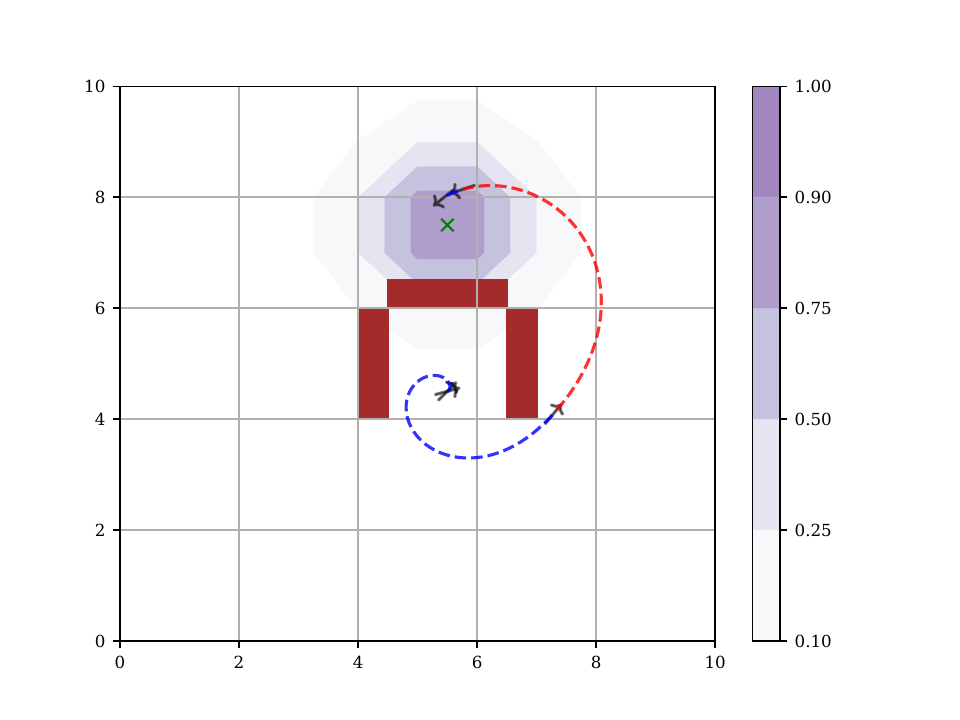}%
     \caption{$A_{\tea}, |\A|{=}2$}
    \end{subfigure}
    \begin{subfigure}[b]{0.25\textwidth}
         \centering
           \includegraphics[width=\linewidth]{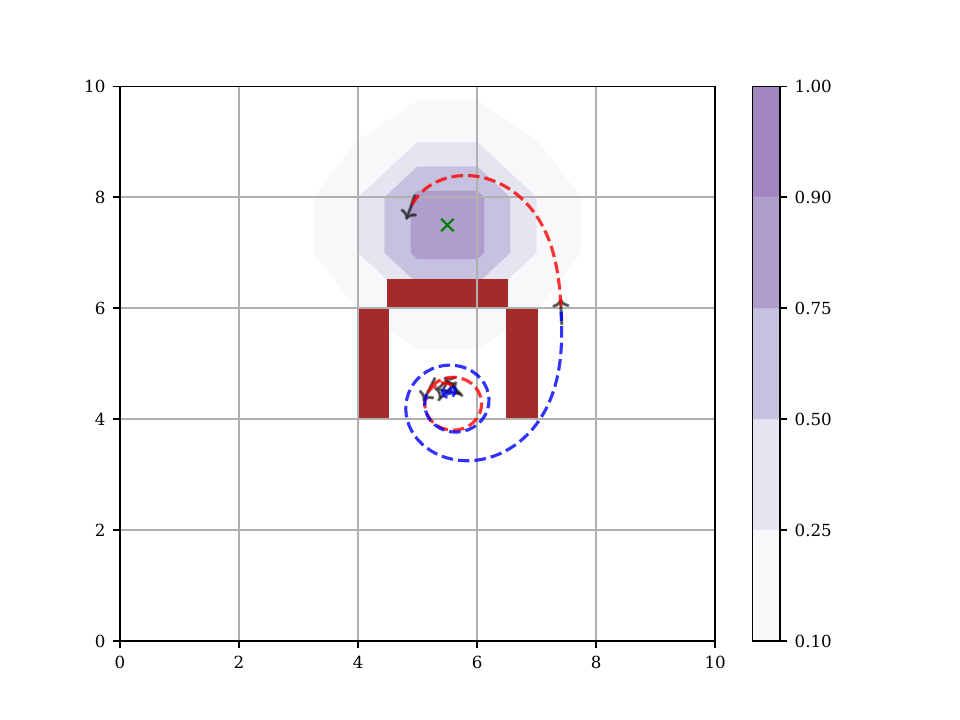}%
     \caption{$A_{\tea}, |\A|{=}102$}
    \end{subfigure}
    \caption{When the number of actions is small (as in a and b), both the agents are able to solve the problem. But when the number of actions increases (c), $A_{\tea}$ is still able to solve the problem while $A_{\naive}$ fails due to large memory requirements.
    The shape of the curves is an artifact of the action space in this environment. Note that a constant action in acceleration space yields curved paths. The path in (c) is composed of 4 actions of different durations, as marked by the colors.
    }%
    \label{fig:exp_u_maps}
\end{figure*}

%% file: sections/experiments/exp_varying_agents_behaviour.tex
\input{tables/varying-gammas/cave-mini-g1-0.99-vary-g2}

\textbf{How does varying the two discount factors impact the agent's behaviour?}
The experiments so far use a single value of $\gamma$ while the proposed formulation in \Cref{eq:proposed_objective} has two discount factors. 
As we discuss next, while some expectations on the effect of $\gamma_1$ and $\gamma_2$ are intuitive, a complete characterization is not obvious.

First, we note that even though $D_{\tea}$ is fixed, the primitive planning horizon is 
dependent on the duration of the actions chosen by the planner. This can result in different levels of discounting in different trajectories. In contrast, in the standard framework, the discounting due to a planning horizon $D_{\naive}$ is fixed. This makes predicting the behavior of $A_{\tea}$, upon varying $\gamma_1$ and $\gamma_2$, difficult.
Second,
for all decision steps $k > 1$, $\gamma_1$ will always be the dominating component in the objective. 
This is because the exponent term for $\gamma_1$ is the number of primitive steps before the decision step $k$, while the exponent term for $\gamma_2$ is proportional to the duration of the temporally-extended action.

To explore the impact of $\gamma_1$ and $\gamma_2$, we fix one and vary the other. %
We use the cave-mini map (\Cref{fig:cave_mini_example}) in the Dubins Car environment, where the agent has to navigate through multiple obstacles.
We have two hypotheses. First, for a fixed $\gamma_1$, smaller values of $\gamma_2$ will 
prefer shorter action durations and hence larger number of decision steps. Second, for a fixed $\gamma_2$, decreasing $\gamma_1$ should 
prefer longer action durations and a smaller total number of primitive actions. The intuition is that if we reduce $\gamma_1$ and the planner does not reduce the number of primitive actions, then the impact of discounting on the overall objective will be higher. \Cref{tab:exp_g1_fixed_g2_vary} confirms both these trends.

%% file: tables/varying-gammas/cave-mini-g1-0.99-vary-g2.tex
\begin{wraptable}{R}{0.55\textwidth}
    \centering
    \scriptsize
 \begin{tabular}{lcccc}
\toprule
& $\gamma_1$ & $\gamma_2$ & Decision Steps & Primitive Steps  \\
\midrule
\multirow{5}{*}{\raisebox{-\heavyrulewidth}{Case 1: Fixed $\gamma_1$}} &
\multirow{5}{*}{\raisebox{-\heavyrulewidth}{0.99}} 
& 1.0 & 20.2 $\pm$ 0.84 & 121.8 $\pm$ 2.95 \\
& & 0.99 & 20.2 $\pm$ 0.45 & 122.0 $\pm$ 2.00\\
& & 0.9 & 21.6 $\pm$ 1.34 & 121.8 $\pm$ 0.84\\
& & 0.8 & 22.2 $\pm$ 0.84 & 122.6 $\pm$ 1.52\\
& & 0.7 & 23.2 $\pm$ 0.45 & 122.8 $\pm$ 1.64\\
 \midrule
\multirow{4}{*}{\raisebox{-\heavyrulewidth}{Case 2: Fixed $\gamma_2$}} &
1.0 & \multirow{4}{*}{\raisebox{-\heavyrulewidth}{1.0}} & 19.0 $\pm$ 0.71 & 128.6 $\pm$ 1.52 \\
& 0.99 &  & 20.2 $\pm$ 0.84 & 121.8 $\pm$ 2.95 \\
& 0.95 &  & 18.6 $\pm$ 0.55 & 117.6 $\pm$ 0.89 \\
& 0.9 &  & 16.4 $\pm$ 0.55 & 115.4 $\pm$ 0.55 \\
\bottomrule
\end{tabular}
\caption{When $\gamma_1$ is fixed, decreasing $\gamma_2$ leads to an increase in number of decision steps. When $\gamma_2$ is fixed, decreasing $\gamma_1$ leads to a decrease in the number of primitive steps. Results averaged across 5 random seeds.}
\label{tab:exp_g1_fixed_g2_vary}
\end{wraptable}

%% file: sections/experiments/exp_online_learning.tex
\input{images/exp_online_learning/main}

\textbf{Do the benefits transfer if we learn the dynamics?}
Having access to the transition and reward function is not realistic. We would like to be able to learn these as we interact with the environment. For this experiment, we use Ant, Half Cheetah, Hopper, Reacher, Pusher and Walker from MuJoCo along with Cartpole. By default, the MuJoCo simulators use a preset value of frame-skip which vary depending on the environment.  This results in the effective timescale being greater than the original timescale. 
For our experiments, we modify the environments to use a frame-skip of 1.

$A_{\tea}$ uses a learned temporally-extended model for planning, while for evaluation we wrap the primitive transition and reward functions in an iterative loop similar to $F_{\text{IP}}$. 
$A_{\tea}(\text{F})$ uses a fixed $\dtmax$ throughout the entire learning process while $A_{\tea}(\text{D})$ uses the proposed MAB framework to dynamically select the value of $\dtmax$ at the start of every episode. 

Each experiment is run for a specific number of iterations, where in each iteration, we train the model using multiple mini-batches drawn from the dataset of past transitions, and then use the trained model to act in the environment for one episode. 
The rewards collected during the episode are used for evaluating the agent's performance. 
We plot the mean and standard deviation, across 5 different seeds, of the running average of scores obtained as well as the number of decisions made by the agents across training iterations. 
The main results from our experiments are shown in \Cref{fig:learning_curves_online_learning}.  

First, we review the results across environments before discussing environment specific observations. Overall, $A_{\tea}$ performs better than $A_{\naive}$ while being faster. $A_{\tea}(\text{F})$, with a suitable selection of $\dtmax$, results in the best performance. 
$A_{\tea}(\text{D})$ requires more training iterations than $A_{\tea}(\text{F})$ as it has to spend time exploring all the candidate arms but it is able to catch up to the performance of $A_{\tea}(\text{F})$ which uses a carefully selected value of $\dtmax$. 
In addition, $A_{\tea}$ is significantly faster than $A_{\naive}$ in terms of computation time (see \Cref{tab:wall_times} in the Appendix).
The speed-up for $A_{\tea}$ comes from three sources - 
\begin{enumerate*} [label=(\roman*)]
\item  using \TEAspace leads to fewer decision points,
\item the reduction in search space, due to \TEAspace further reduces planning time and,
\item a decrease in decision points results in a smaller sized dataset of past transitions that the model needs to learn from, thus reducing the model training time.
\end{enumerate*}

In Cartpole, an episode lasts for 200 primitive actions, unless the pole falls which terminates the episode.
By default, the agent gets a reward of 1 at each timestep. We find that this reward formulation leads to highly unstable learning using the standard framework. The spike in number of decision points for $A_{\naive}$ ($D_{\naive}=30$) is because the agent learns to keep the pole from falling for the entire duration of the episode, but it soons crashes and the agent is unable to recover. Following \cite{wang2019benchmarking}, we also experiment with a more informative reward and find that it stabilizes the learning of the standard framework. Both variants of $A_{\tea} (D_{\tea}=5)$ work well with both the reward functions.  

An episode in Ant, Half Cheetah, Hopper and Walker lasts for 1000 primitive actions. By default, the reward per timestep in these environments is not bounded. As our bandit framework uses a UCB heuristic for selecting $\dtmax$, we modify the reward per timestep to be between 0 and 1. Details of the reward functions are in \Cref{sec:appendix_env_overview}.   
In Half Cheetah, $A_{\tea} (D_{\tea} = 15)$ performs similarly to $A_{\naive} (D_{\naive}=90$) but is 8x faster. In Ant, Hopper and Walker, $A_{\tea}$ significantly outperforms $A_{\naive}$. $A_{\tea}(\text{F})$ results in the best performance among agents showing the importance of properly selecting $\dtmax$. $A_{\tea}(\text{D})$ explores all the possible candidate arms before fixating on one, as can be seen from the plots of number of decision points. 
The value of $\dtmax$ eventually chosen by $A_{\tea}(\text{D})$ is close to $A_{\tea}(\text{F})$. Note that $A_{\tea}(\text{D})$ does not have the flexibility to choose a value of $\dtmax$ arbitrarily but has to choose from a fixed list of exponentially spaced $\dtmax$ candidates. 

In Pusher and Reacher, the performance of $A_{\tea}$ is again similar to $A_{\naive}$. But in this case, the number of decision points for $A_{\tea}$ is also similar to that of $A_{\naive}$, indicating that $A_{\tea}$ identifies that a small decision timescale works better here. This illustrates that in cases when a long duration is not suitable, $A_{\tea}$'s performance is still competitive with the standard solution.

\textbf{How do temporally-extended actions compare to action repeats?} Action repeats are a special case of temporally-extended actions where all actions have the same fixed duration. While using temporally-extended actions provides flexibility to an agent in terms of choice of duration for each temporally-extended action, it does add a slight overhead when compared to action repeats. In order to understand whether the agent makes use of the provided flexibility or simply falls back to action repeats, we examine the distribution of action duration chosen by $A_{\tea}(\text{F})$ in Ant, Half Cheetah, Hopper, and Walker \emph{during the course of one episode}. \Cref{fig:histogram_of_action_repeats} shows the histogram of discretized action durations as action repeats chosen by the agent using the model at the end of training. 

While in Ant and Half Cheetah, $A_{\tea}(\text{F})$ mostly chooses actions with similar $\delta t$, in Hopper and Walker, the values of $\delta t$ vary significantly during the episode. This indicates that in environments like Ant and Half Cheetah, the agent can obtain a similar performance by using a fixed $\delta t$ throughout the episode. However, in environments like Hopper and Walker, the agent benefits from flexibly adapting action duration in each step of execution.
\input{images/histogram_of_action_repeats/main}

%% file: images/exp_online_learning/main.tex
\begin{figure*}[tbp]%
    \centering%
    \begin{subfigure}[b]{0.245\textwidth}
         \centering
           \includegraphics[width=\linewidth]{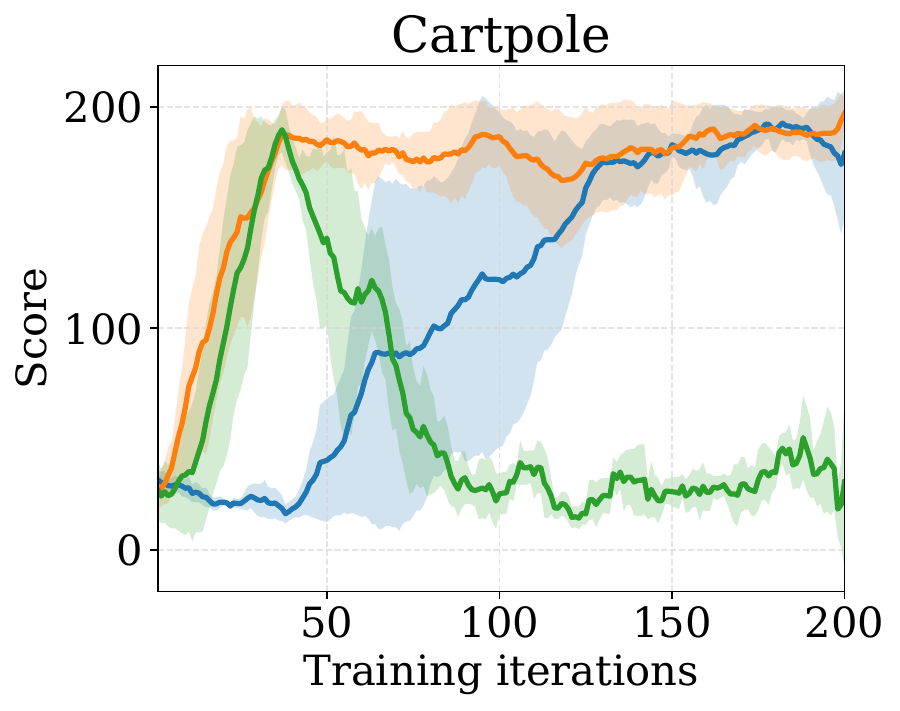}%
    \end{subfigure}
    \hfill
    \begin{subfigure}[b]{0.245\textwidth}
         \centering
           \includegraphics[width=\linewidth]{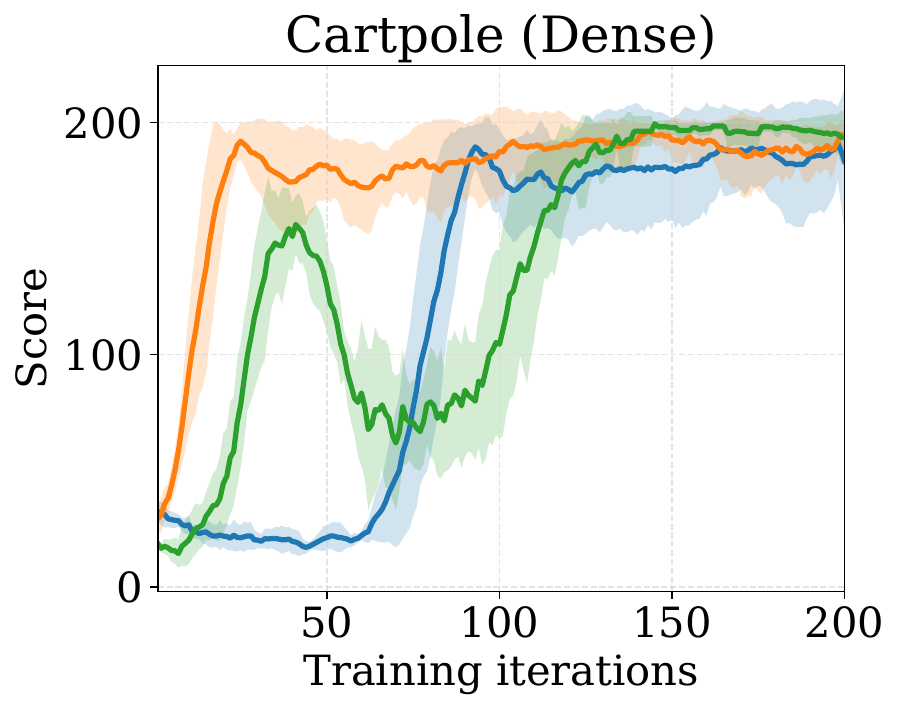}%
    \end{subfigure}
    \hfill
    \begin{subfigure}[b]{0.245\textwidth}
         \centering
           \includegraphics[width=\linewidth]{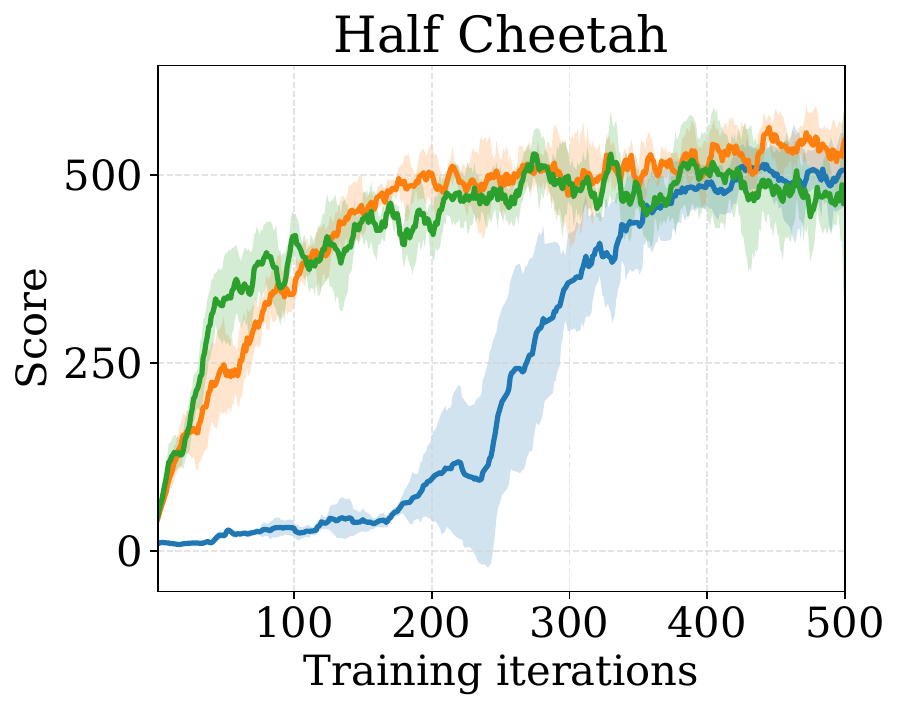}%
    \end{subfigure}
    \hfill
    \begin{subfigure}[b]{0.245\textwidth}
         \centering
           \includegraphics[width=\linewidth]
           {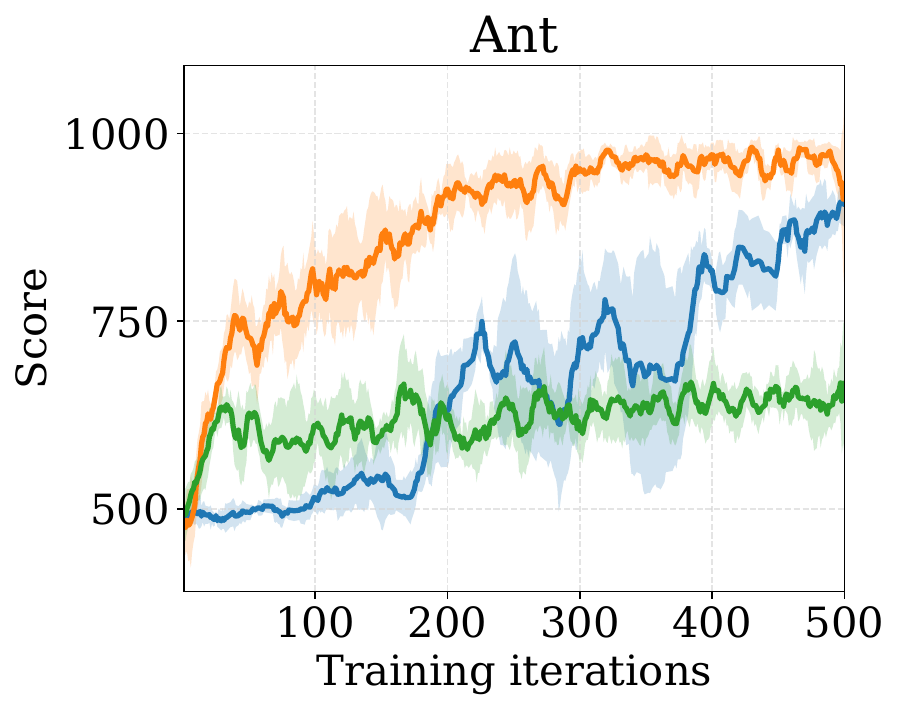}%
    \end{subfigure}
    \\
    \begin{subfigure}[b]{0.245\textwidth}
         \centering
         \includegraphics[width=\linewidth]{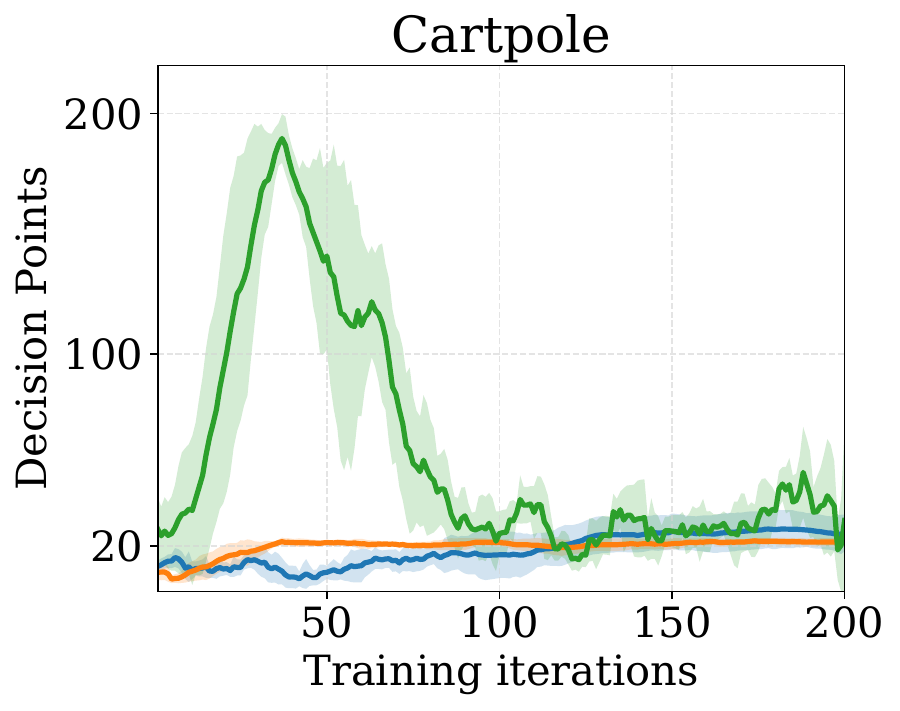}%
    \end{subfigure}
    \begin{subfigure}[b]{0.245\textwidth}
         \centering
         \includegraphics[width=\linewidth]{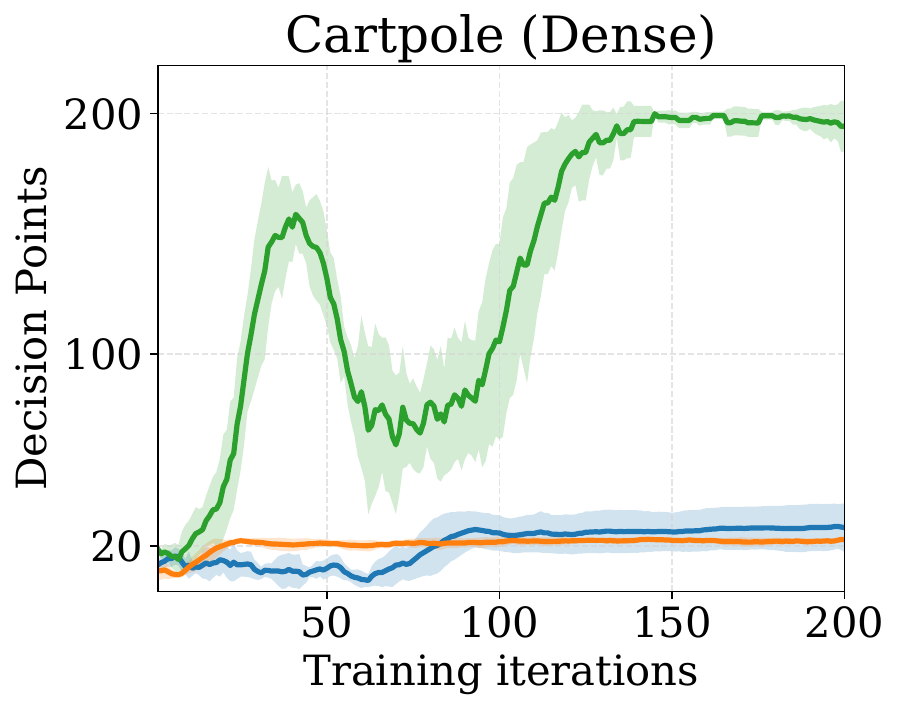}%
    \end{subfigure}
    \hfill
      \begin{subfigure}[b]{0.245\textwidth}
         \centering
         \includegraphics[width=\linewidth]{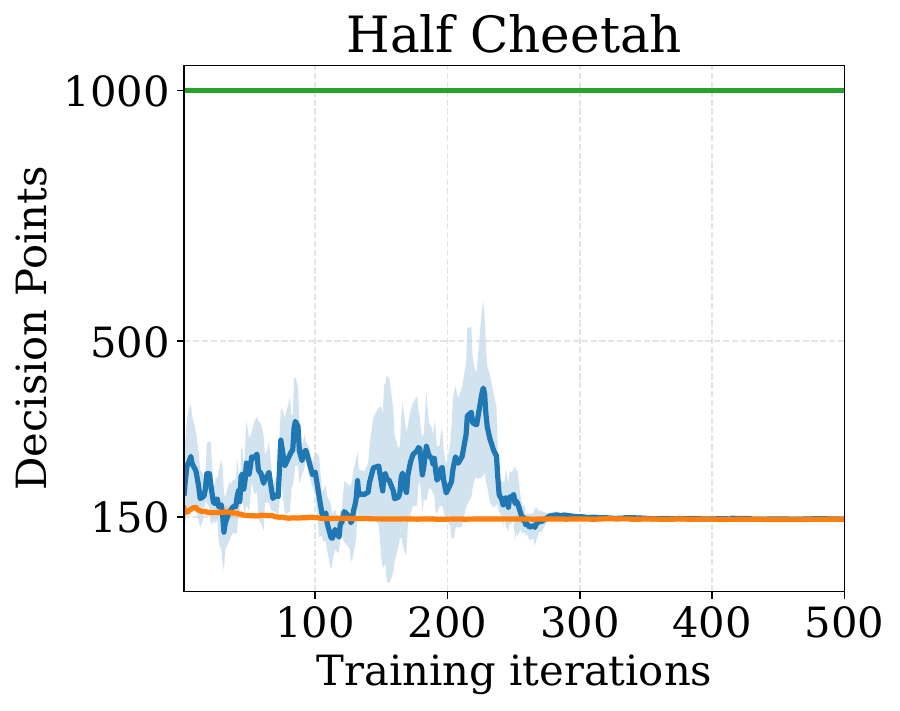}%
    \end{subfigure}
    \hfill
    \begin{subfigure}[b]{0.245\textwidth}
         \centering
         \includegraphics[width=\linewidth]{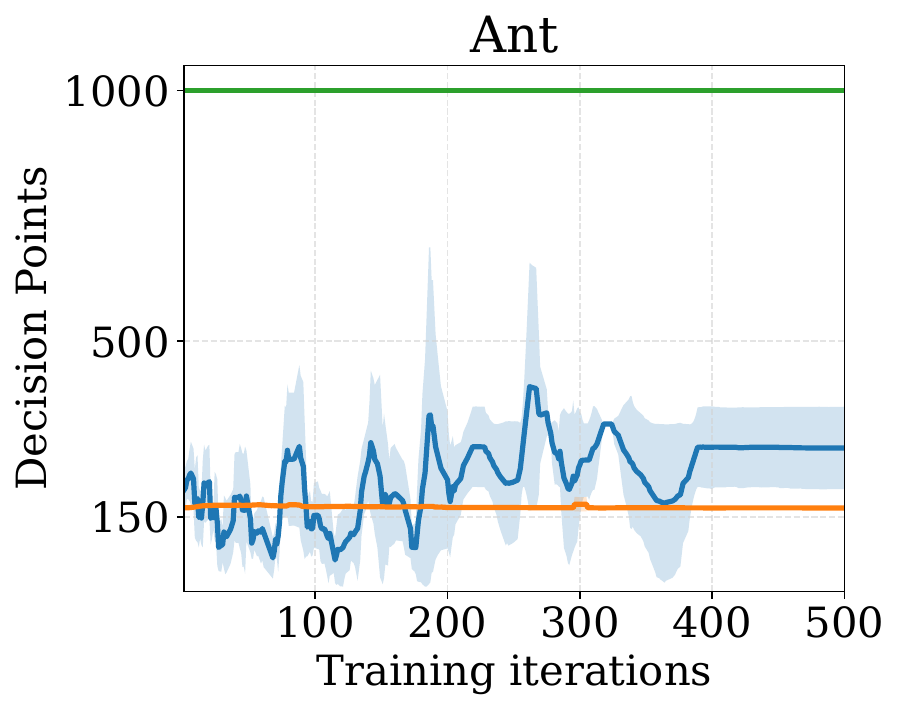}%
    \end{subfigure}
    \\
    \textcolor{lightgray}{\rule{\textwidth}{0.1pt}}
    \begin{subfigure}[b]{0.245\textwidth}
         \centering
           \includegraphics[width=\linewidth]
           {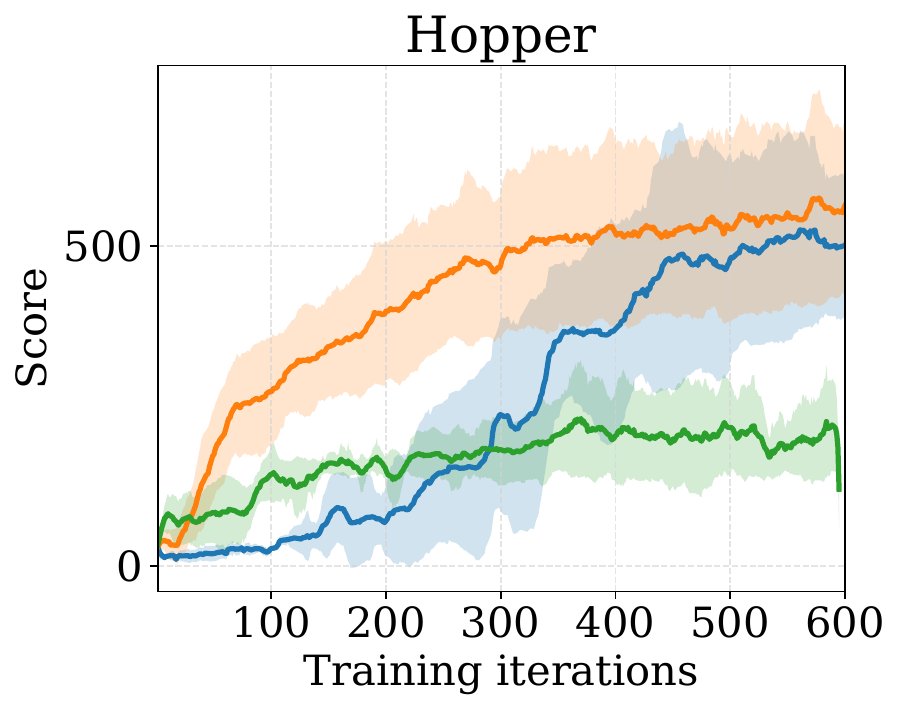}%
    \end{subfigure}
      \begin{subfigure}[b]{0.245\textwidth}
         \centering
           \includegraphics[width=\linewidth]
           {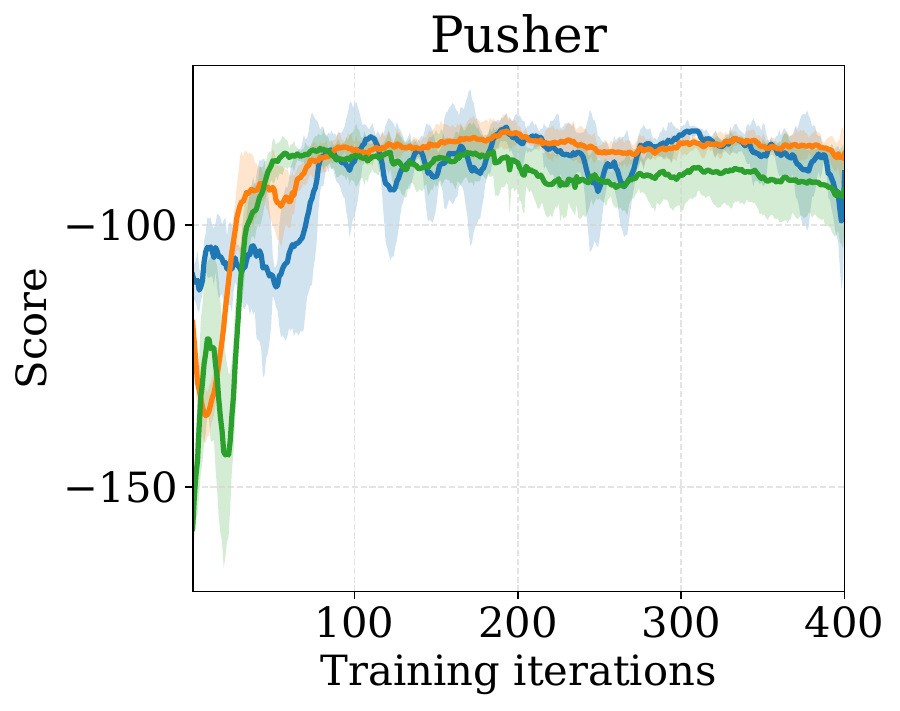}%
    \end{subfigure}
          \begin{subfigure}[b]{0.245\textwidth}
         \centering
           \includegraphics[width=\linewidth]
           {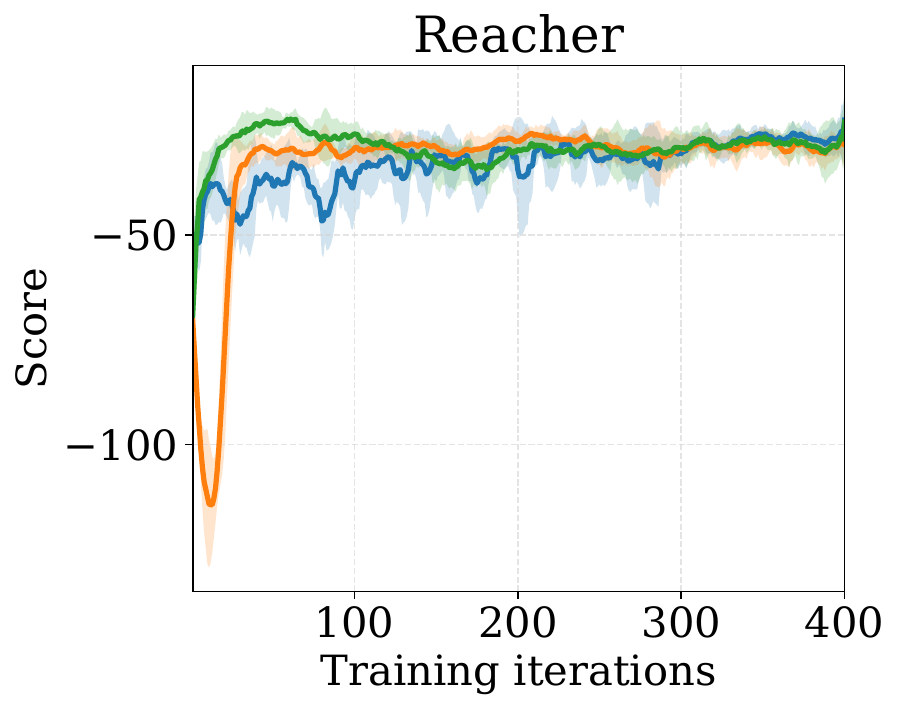}%
    \end{subfigure}
          \begin{subfigure}[b]{0.245\textwidth}
         \centering
           \includegraphics[width=\linewidth]
           {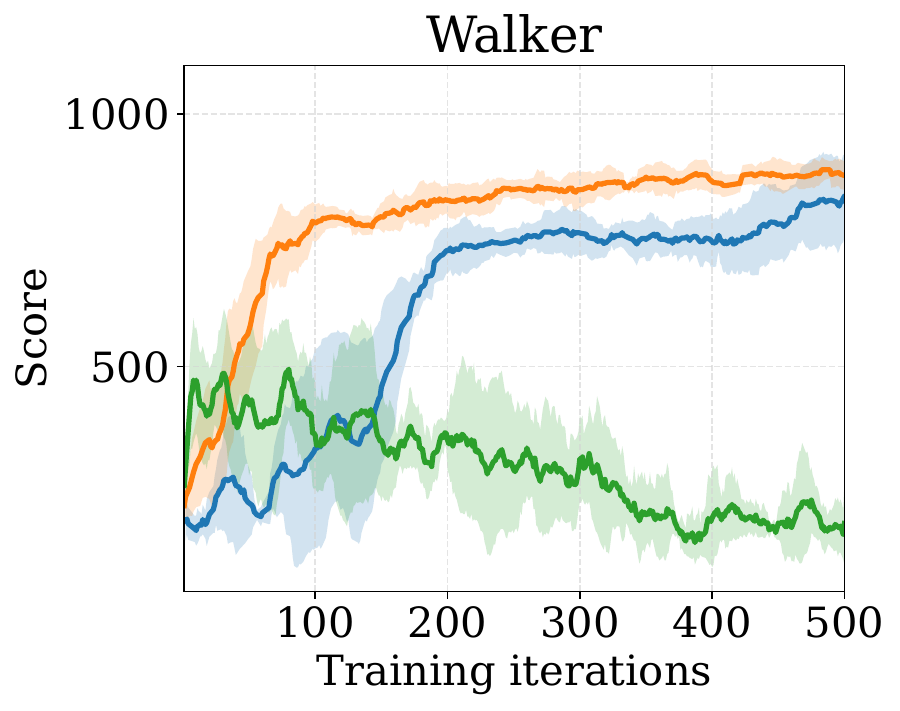}%
    \end{subfigure}
    \\
     \begin{subfigure}[b]{0.245\textwidth}
         \centering
         \includegraphics[width=\linewidth]{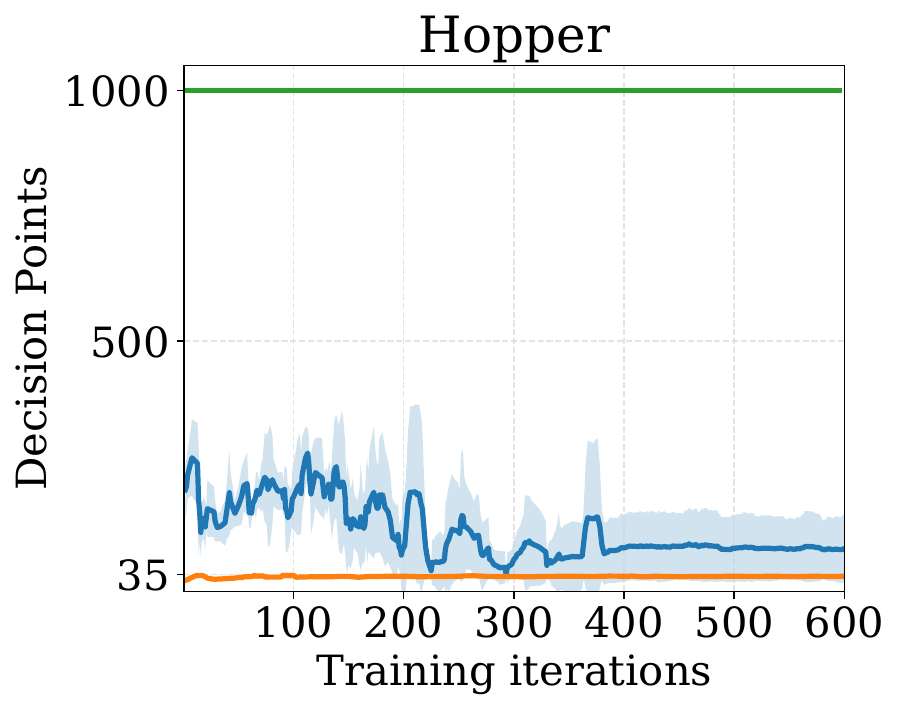}%
    \end{subfigure} 
    \begin{subfigure}[b]{0.245\textwidth}
         \centering
           \includegraphics[width=\linewidth]{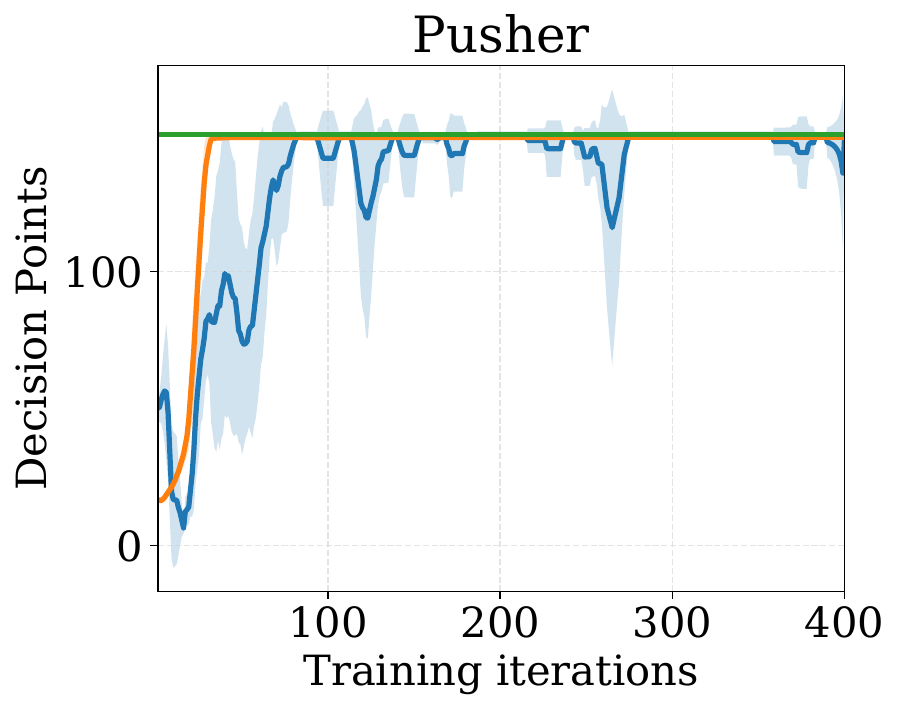}%
    \end{subfigure}
    \begin{subfigure}[b]{0.245\textwidth}
         \centering
           \includegraphics[width=\linewidth]{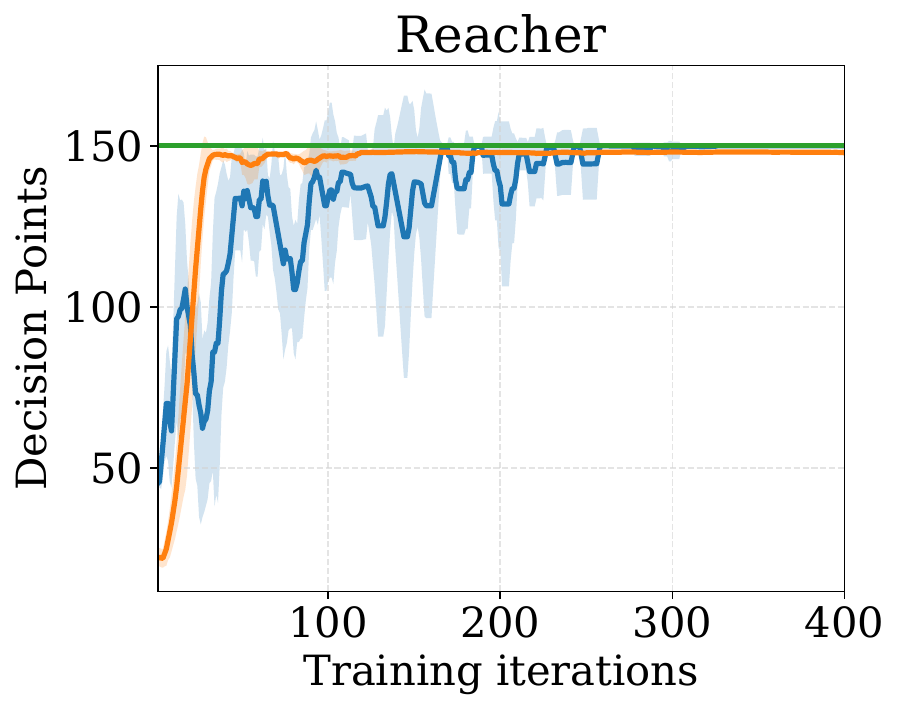}%
    \end{subfigure}
    \begin{subfigure}[b]{0.245\textwidth}
         \centering
           \includegraphics[width=\linewidth]{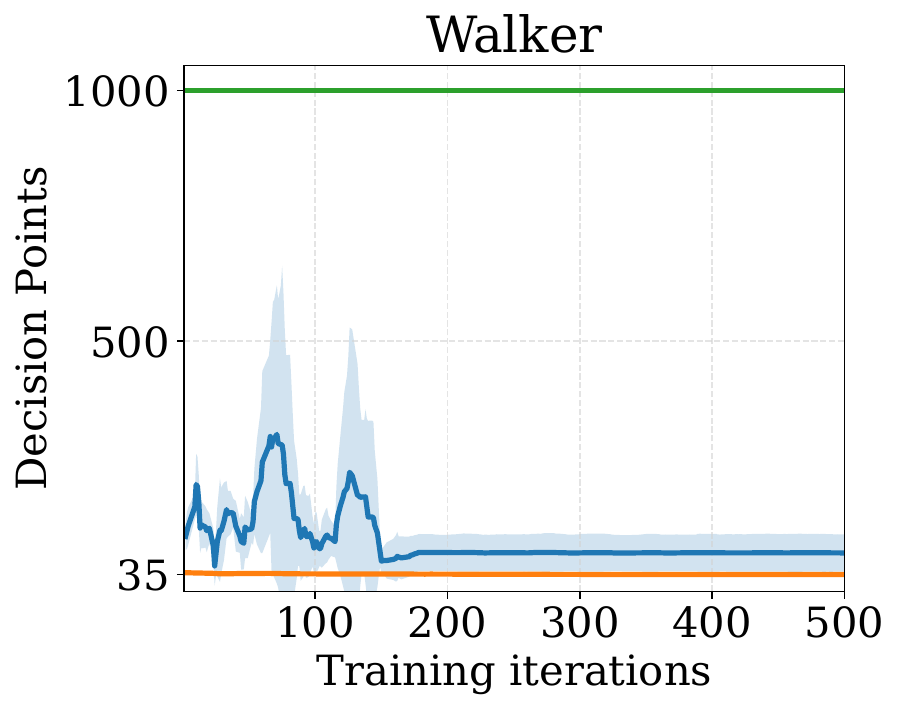}%
    \end{subfigure}
    \\
    \begin{subfigure}[b]{0.33\textwidth}
         \centering
         \includegraphics[width=\linewidth]{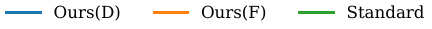}%
    \end{subfigure}
    
    \caption{Mean and standard deviation of the running averages (window size=10) of scores and number of decisions taken by the agents in classical control and MuJoCo domains across 5 different seeds. \textcolor{ablue}{Ours(D)} selects $\dtmax$ automatically using the proposed multi-armed bandit framework while \textcolor{aorange}{Ours(F)} uses fixed $\dtmax$ for every episode. Using temporally-extended actions results in better performance in many environments while requiring fewer decision points.}
    \label{fig:learning_curves_online_learning}
\end{figure*}

%% file: images/histogram_of_action_repeats/main.tex
\begin{figure*}[h]%
    \centering%
    \begin{subfigure}[b]{0.24\textwidth}
         \centering
           \includegraphics[width=\linewidth]{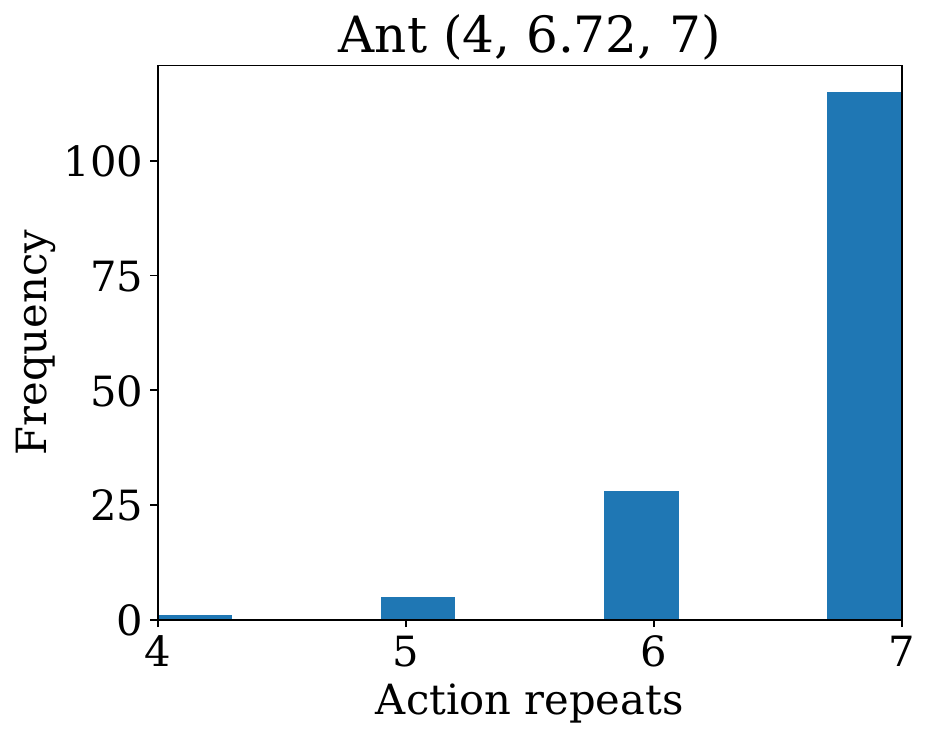}%
    \end{subfigure}
    \hfill
    \begin{subfigure}[b]{0.24\textwidth}
         \centering
           \includegraphics[width=\linewidth]{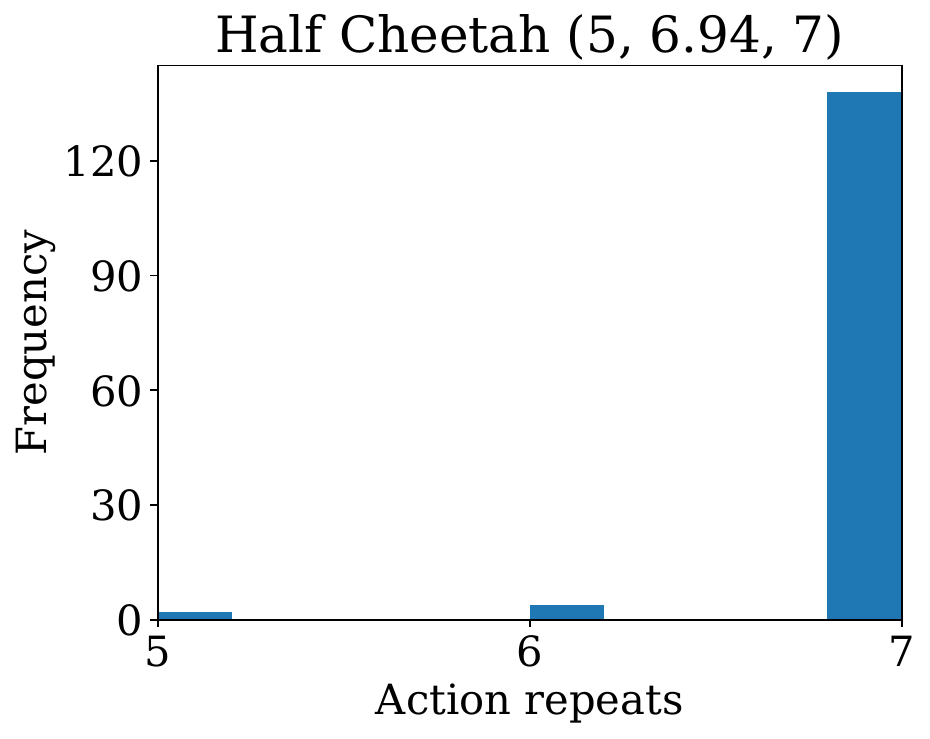}%
    \end{subfigure}
    \hfill
    \begin{subfigure}[b]{0.24\textwidth}
         \centering
           \includegraphics[width=\linewidth]{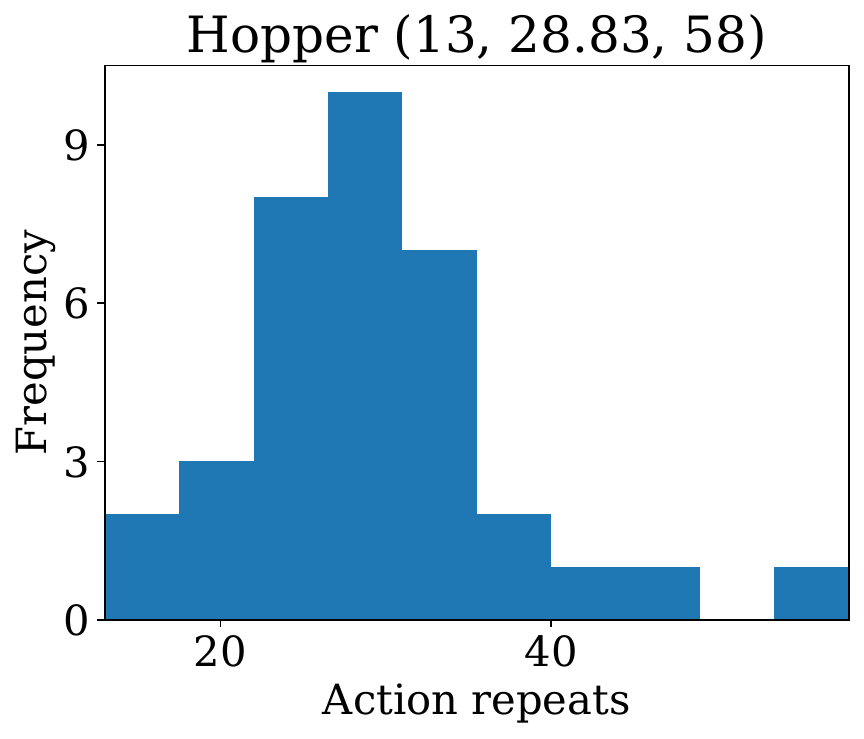}%
    \end{subfigure}
    \hfill    
    \begin{subfigure}[b]{0.24\textwidth}
         \centering
           \includegraphics[width=\linewidth]{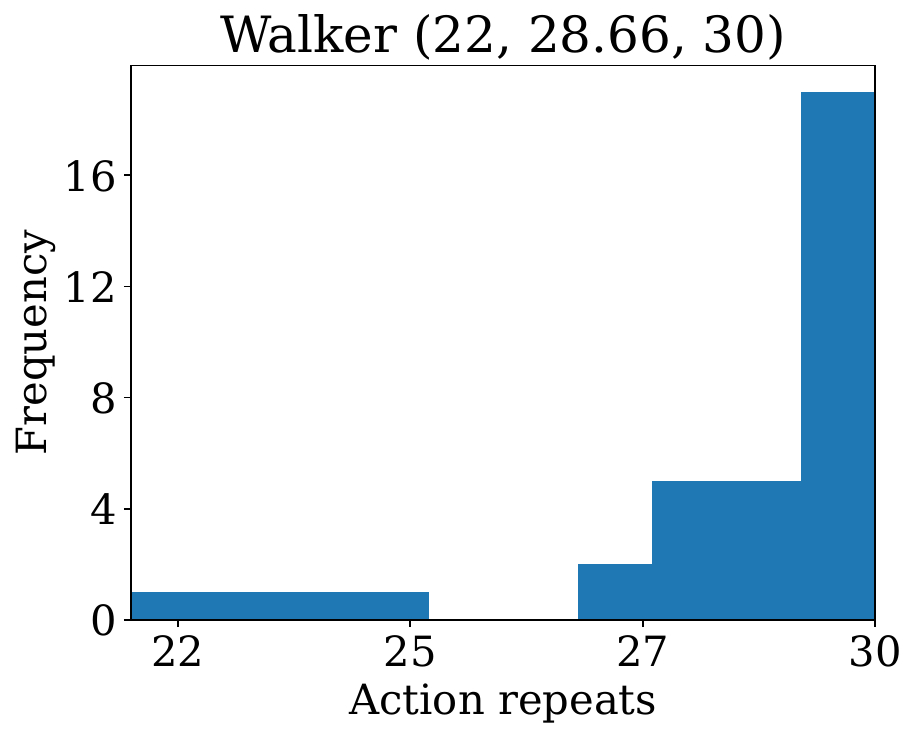}%
    \end{subfigure}
    \caption{Histogram of action duration in terms of primitive action repeats from the one episode at the end of training for Ant, Half Cheetah, Hopper and Walker. In contrast to using a fixed frame-skip, using \TEAs~ allows the agent to take actions of varying durations. Numbers in the title indicate the minimum, average and maximum of action repeats taken.}
    \label{fig:histogram_of_action_repeats}
\end{figure*}

%% file: sections/conclusion/main.tex
\section{Conclusion and Future Work}
Using the standard framework of planning with primitive actions provides more granular control but can be computationally expensive. 
When $\dtenv$ is small, an agent requires a large planning horizon which increases the search space as well as the number of variables it needs to optimize. 
We propose to use \TE~actions where the planner treats the action duration as an additional optimization variable. 
This decreases the complexity of the search by restricting the search space and reducing the planning horizon, which in turn results in a smaller number of variables for the planner to optimize.
Further, we show that, with a suitable $\dtmax$, learning \TE~dynamics models using MBRL can be faster while outperforming the standard framework in many environments. 
Finally, rather than setting $\dtmax$ as a hyperparameter, a MAB framework can be used to dynamically select it. This is slower to converge but provides the same performance benefits as a carefully chosen value of $\dtmax$. 
Our proposed approach does have several potential limitations. First, using \TE~actions does not guarantee improvement in every situation since the generated trajectories are not as flexible as those due to primitive actions. 
Second, the proposed MAB formulation requires fixed $D_{\tea}$ for every bandit arm. Dynamic selection of both $\dtmax$ and $D_{\tea}$ together is left as future work.

%% file: sections/ack.tex
\section*{Acknowledgements}

We would like to thank the reviewers for their time and valueable feedback. This work was partly supported by NSF under grant 2246261. Some of the experiments in this paper were run on the Big Red computing system at Indiana University, supported in part by Lilly Endowment, Inc., through its support for the Indiana University Pervasive Technology Institute.

%% file: sections/checklist.tex
\section*{NeurIPS Paper Checklist}

\begin{enumerate}

\item {\bf Claims}
    \item[] Question: Do the main claims made in the abstract and introduction accurately reflect the paper's contributions and scope?
    \item[] Answer: \answerYes{} %
    \item[] Justification: \emph{In the abstract, we claim that using temporally-extended actions adds additional structure to the search and can thus improve planning and learning using MBRL. In the paper, we showcase this empirically by performing extensive experiments.}

\item {\bf Limitations}
    \item[] Question: Does the paper discuss the limitations of the work performed by the authors?
    \item[] Answer: \answerYes{} %
    \item[] Justification: \emph{In the conclusion, we acknowledge that our method has several potential limitations. The proposed framework does not guarantee improvement in every domain as also noted in MBRL experiments with Reacher and Pusher. Secondly, the meta algorithm for selecting $\dtmax$ assumes that planning horizon is fixed. A more robust solution will select more $\dtmax$ and planning horizon simultaneously.}

\item {\bf Theory assumptions and proofs}
    \item[] Question: For each theoretical result, does the paper provide the full set of assumptions and a complete (and correct) proof?
    \item[] Answer: \answerNA{} %
    \item[] Justification: \emph{There are no theoretical results in the paper.}

    \item {\bf Experimental result reproducibility}
    \item[] Question: Does the paper fully disclose all the information needed to reproduce the main experimental results of the paper to the extent that it affects the main claims and/or conclusions of the paper (regardless of whether the code and data are provided or not)?
    \item[] Answer: \answerYes{} %
    \item[] Justification: \emph{The environment specifications along with the necessary hyperparameters are described in the Appendix.}

\item {\bf Open access to data and code}
    \item[] Question: Does the paper provide open access to the data and code, with sufficient instructions to faithfully reproduce the main experimental results, as described in supplemental material?
    \item[] Answer: \answerYes{} %
    \item[] Justification: \emph{The code can be found at \url{https://github.com/pecey/MBRL-with-TEA/}}

\item {\bf Experimental setting/details}
    \item[] Question: Does the paper specify all the training and test details (e.g., data splits, hyperparameters, how they were chosen, type of optimizer, etc.) necessary to understand the results?
    \item[] Answer: \answerYes{} %
    \item[] Justification: \emph{We provide the relevant experimental configuration in \Cref{sec:hyperparameters}}

\item {\bf Experiment statistical significance}
    \item[] Question: Does the paper report error bars suitably and correctly defined or other appropriate information about the statistical significance of the experiments?
    \item[] Answer: \answerYes{} %
    \item[] Justification: \emph{We report the mean and the standard deviation across 5 seeds for each experiment.}

\item {\bf Experiments compute resources}
    \item[] Question: For each experiment, does the paper provide sufficient information on the computer resources (type of compute workers, memory, time of execution) needed to reproduce the experiments?
    \item[] Answer: \answerYes{} %
    \item[] Justification: \emph{We provide the details regarding the computational resources in \Cref{sec:computational_resources}}
    
\item {\bf Code of ethics}
    \item[] Question: Does the research conducted in the paper conform, in every respect, with the NeurIPS Code of Ethics \url{https://neurips.cc/public/EthicsGuidelines}?
    \item[] Answer: \answerYes{} %
    \item[] Justification: \emph{Our research conforms to the NeurIPS Code of Ethics.}

\item {\bf Broader impacts}
    \item[] Question: Does the paper discuss both potential positive societal impacts and negative societal impacts of the work performed?
    \item[] Answer: \answerNA{} %
    \item[] Justification: \emph{The paper presents foundational research that is not tied to particular application. Broader impact concerns are hence generic.} %
    
\item {\bf Safeguards}
    \item[] Question: Does the paper describe safeguards that have been put in place for responsible release of data or models that have a high risk for misuse (e.g., pretrained language models, image generators, or scraped datasets)?
    \item[] Answer: \answerNA{}%
    \item[] Justification: \emph{We are not releasing any data or models as a part of this research.}

\item {\bf Licenses for existing assets}
    \item[] Question: Are the creators or original owners of assets (e.g., code, data, models), used in the paper, properly credited and are the license and terms of use explicitly mentioned and properly respected?
    \item[] Answer: \answerYes{} %
    \item[] Justification: \emph{All the code required for this research was written by using standard open source frameworks.}

\item {\bf New assets}
    \item[] Question: Are new assets introduced in the paper well documented and is the documentation provided alongside the assets?
    \item[] Answer: \answerNA{} %
    \item[] Justification: \emph{We are not releasing any new assets.}

\item {\bf Crowdsourcing and research with human subjects}
    \item[] Question: For crowdsourcing experiments and research with human subjects, does the paper include the full text of instructions given to participants and screenshots, if applicable, as well as details about compensation (if any)? 
    \item[] Answer: \answerNA{} %
    \item[] Justification: \emph{This work does not involve crowdsourcing or research with human subjects.}

\item {\bf Institutional review board (IRB) approvals or equivalent for research with human subjects}
    \item[] Question: Does the paper describe potential risks incurred by study participants, whether such risks were disclosed to the subjects, and whether Institutional Review Board (IRB) approvals (or an equivalent approval/review based on the requirements of your country or institution) were obtained?
    \item[] Answer: \answerNA{} %
    \item[] Justification: \emph{This work does not involve crowdsourcing or research with human subjects.}

\item {\bf Declaration of LLM usage}
    \item[] Question: Does the paper describe the usage of LLMs if it is an important, original, or non-standard component of the core methods in this research? Note that if the LLM is used only for writing, editing, or formatting purposes and does not impact the core methodology, scientific rigorousness, or originality of the research, declaration is not required.
    \item[] Answer: \answerNA{} %
    \item[] Justification: \emph{The paper does not use LLMs in any form.}

\end{enumerate}

%% file: appendix/main.tex
\section*{\LARGE Appendix}
\label{sec:appendix}

\vspace*{30pt}
\section*{Table of Contents}
\vspace*{-5pt}
\startcontents[sections]
\printcontents[sections]{l}{1}{\setcounter{tocdepth}{2}}

\clearpage

\renewcommand{\thefigure}{A\arabic{figure}}
\setcounter{figure}{0}

\renewcommand{\thetable}{A\arabic{table}}
\setcounter{table}{0}

\renewcommand{\thealgorithm}{A\arabic{algorithm}}
\renewcommand{\theHalgorithm}{A\arabic{algorithm}}
\setcounter{algorithm}{0}

\input{appendix/list_of_symbols.tex}

\input{appendix/returns_due_to_tea}

\input{appendix/algorithms/iterative_primitive_transition_function}

\input{appendix/env_details}
\input{appendix/hyperparams}

\input{appendix/computational_resources}
\input{appendix/additional_planning_experiments}

\input{appendix/additional_mbrl_results}

\input{appendix/ablation/main}

%% file: appendix/list_of_symbols.tex
\section{List of Symbols}

\input{tables/appendix/list_of_symbols.tex}

%% file: tables/appendix/list_of_symbols.tex
\begin{table*}[tbhp]
\centering
\begin{tabular}{p{0.15\linewidth} p{0.75\linewidth}}
\toprule
$\delta_t$ & Base timescale of the system.\\
$\dtenv$ & Decision timescale of the environment = Frameskip $\times$ $\delta_t$. \\
$\delta t$ & Action duration chosen by the planner. \\
$\delta t_{\min}, \delta t_{\max}$ & Range for the action duration. \\ 
$f$ & One-step dynamics function. $\Smdp \times \A \rightarrow \Smdp$ \\
$\fip$ & Iterative primitive transition function. \\
$F$ & Temporally extended dynamics function. $\Smdp \times (\A \cup \dt) \rightarrow \Smdp$\\
$\fhatip$ & Indirectly approximates $F$. Learns a one-step model to approximate $f$ and then iterates over it. \\
$\hat{F}_{\text{TE}}$ & Directly approximates $F$. Learns a model to predict the next state due to a temporally-extended action.\\
$A_{\text{STD}}$ & TS$\infty$ variant of the CEM agent from PETS \citep{chua2018deep}. \\
$A_{\tea}(\text{F})$ & Proposed agent that used a fixed and manually chosen value of $\delta t_{\max}$. \\
$A_{\tea}(\text{D})$ & Proposed agent that used a MAB framework to dynamically select $\delta t_{\max}$ before every episode. \\
$D_{\text{STD}}$ & Planning horizon used by the standard agent. \\
$D_{\tea}$ & Planning horizon used by the proposed agent. \\
\end{tabular}
\end{table*}

%% file: appendix/returns_due_to_tea.tex
\section{Connection to \cite{ni2022continuous}}\label{sec:connection_to_continous_control_on_time}

{\citet{ni2022continuous} propose to learn a policy that outputs the action variables as well as the time scale. They have a similar looking objective. In this section, we compare our objective to theirs. 
First, we repeat \Cref{eq:proposed_objective_j2} which computes the returns of a trajectory $\tau$ using the proposed framework:
\begin{align*}
    J(\tau) = \sum_{k = 1}^{L(\tau)} \gamma^{e_{<k}} \sum_{t = 1}^{e_k} \gamma^{t-1} \R_{(e_{<k} + t)} .
\end{align*}

A similar objective is used by \cite{ni2022continuous}. However, it is important to note that they control the timescale while we keep the timescale fixed and control the action duration. Let $\delta_k \in [\delta_{\min}, \delta_{\max}]$ be the timescale associated with the action at decision step $k$. For simplicity, they assume that the timescale is in integers (in physical unit of seconds), which means that $\delta_k$ is also the number of execution steps associated with the action. Using this, their objective reduces to
\begin{align}
    J(\tau) = \sum_{k=1}^{L(\tau)} \gamma^{e_{<k}} \R(s_{e_{<k} + 1}, a_{k}) \delta_k  
\end{align}
where $\R(s_{e_{<k} + 1}, a_{k}) \delta_k$ is a linear approximation of the reward function. So, the objective used by \citet{ni2022continuous} can be viewed as an approximation of \Cref{eq:proposed_objective_j2} where the reward function due to a temporally-extended action has been replaced by a linear approximation.

%% file: appendix/algorithms/iterative_primitive_transition_function.tex
\input{algorithm/iterative_primitive_transition_fn}

\section{Iterative Primitive Dynamics Function}
In most cases we have access to the primitive dynamics function $(f)$. 
A trivial way of making $f$ work with \TEAs~ is to wrap it in a loop. 
For the computation to be exact, $f$ should be explicitly dependent on time. 
If it is implicit, performing the computation at line 5 %
of \Cref{alg:long_horizon_transition_function} will not be possible and the algorithm computes the number of action-repeats instead. 

The reward is simply aggregated in the loop and is not shown explicitly here. Note that for a given environment, $f$ is fixed, while $s_k$, $a_k$ and $\dtk$ are dependent on the timestep.

%% file: algorithm/iterative_primitive_transition_fn.tex
\begin{algorithm}[b]
\caption{$\fip$ : Iterative primitive dynamics function}\label{alg:long_horizon_transition_function}
\begin{algorithmic}[1]
\Require primitive dynamics function $(f)$, current state $(s_k)$, \newline 
\hspace*{2.2em} \TEA {} $(a_k)$, duration of action $(\dtk)$
\State $\text{repeats} = \lfloor{\nicefrac{\dtk}{\dtenv}}\rfloor$
\For{\text{$i = 1$ to repeats}}
\State $s_k = f(s_k, a_k, \dtenv)$ \Comment{Control $n$}
\EndFor
\State $s_{k+1} = f(s_k, a_k, \dtk ~ \texttt{mod} ~ \dtenv)$  \Comment{Control $\delta_t$}
\State \Return $s_{k+1}$
\end{algorithmic}
\end{algorithm}

%% file: appendix/env_details.tex
\section{Environment Overview}
\label{sec:appendix_env_overview}

In this section, we provide an overview of the environments used in our experiments. \Cref{tab:env_reward_fns} lists the dimensionality of the observation and the action space of the environments, the reward functions along with the episode length (which is equivalent to the maximum number of primitive steps allowed).

\input{tables/appendix/reward_fns}

\textbf{Details for Dubins Car environment : } We use the Dubins Car environment of \citet{chatterjee2023disprod}. The state space comprises of the $x$ and $y$ co-ordinates, and the orientation $(\theta)$ of the car, along with the current linear $(v)$ and angular velocity $(\omega)$ of the car and the action space is the change in linear velocity $(\Delta v)$ and angular velocity $(\Delta \omega)$. An episode terminates if the agent reaches the goal or if it takes 300 primitive steps.

\subsection{MuJoCo environments}

\textbf{Controlling $\dtenv$ :}
Each environment defined in MuJoCo has its own value of $\delta_t$ which is defined in the corresponding XML file along with the integrator to be used. Further, each environment has a predefined frame-skip which is defined in the constructor of the corresponding class. The combination of these two control $\dtenv$ in MuJoCo. In order to operate at the base $\delta_t$, we simply set the frame-skip value to 1.

\textbf{Reward functions :} 
The default reward functions in MuJoCo are not bounded. For the MAB framework, we require a bounded reward function. The problem is amplifed in Ant, Half Cheetah, Hopper and Walker, where the rewards over an episode can be very large. Therefore, for these environments, we adapt the bounded reward functions from DeepMind Control Suite \citep{dmcontrol}, so that the agents always get a reward between 0 and 1, with the specifics depending on the environment.

\textbf{Maximum episode lengths :} MuJoCo environments have a predefined value of frame-skip. The default episode length is with respect to this frame-skip. For example, Half Cheetah has a frame-skip of 5 and a episode length of 1000. An equivalent episode in Half Cheetah with a frame-skip of 1 should be of length 5000. However, we limit the episode to 1000 as otherwise $A_{\naive}$ would have taken too long to train. Even with episode lengths of 1000, $A_{\naive}$ requires more than 40 hours to train (\Cref{tab:wall_times}). 

%% file: tables/appendix/reward_fns.tex
\begin{table}[tbph]
    \centering
    \scriptsize
    \begin{tabular}{lcccl}
    \toprule
        Domain & nS & nA & Episode steps & Reward Function\\
    \midrule
Dubins Car                  & 5 & 2 & 300 & $100 \times \mathds{1}_{\text{distance from goal} < 0.5} - 10 \times \mathds{1}_{\text{collision = True}}$ \\
Cartpole                    & 4 & 1 & 200 & $\mathds{1}_{x,\theta \text{ within bounds}}$\\
Cartpole (Dense)            & 4 & 1 & 200 & $\cos(\theta_t) - 0.001 x_t^2$\\
Mountain Car                & 2 & 1 & 200 & $\mathds{1}_{x_t = \text{goal}} \times 100 - 0.1 a_t^2$\\
IPC Mountain Car            & 2 & 1 & 500 & $\mathds{1}_{x_t = \text{goal}} \times 100 - 0.1 a_t^2$\\
Reacher                     & 17 & 7 & 150 & $-\text{dist(finger, obj)} - a_t^2$\\
Pusher                      & 23 & 7 & 150 & $ -0.5 \times \text{dist(finger, obj)} - \text{dist(object, goal)} - 0.1a_t^2$\\
Ant                         & 27 & 8 & 1000 & $\text{sigmoid}(\dot{x_t}) \text{ if } \dot{x_t} \le 0.5 \text{ else } 1$\\
Half Cheetah                & 18 & 6 & 1000 & $\text{sigmoid}(\dot{x_t}) \text{ if } \dot{x_t} \le 10 \text{ else } 1$\\
Hopper                      & 11 & 3 & 1000 & $(\text{sigmoid}(\dot{x_t}) \text{ if } \dot{x_t} \le 2 \text{ else } 1) \times \text{healthy reward}$\\
Walker                      & 17 & 6 & 1000 & $\nicefrac{1}{2} ((\text{sigmoid}(\dot{x_t}) \text{ if } \dot{x_t} \le 1 \text{ else } 1) + \text{healthy reward})$\\
\bottomrule
\smallskip
\end{tabular}
\caption{State and action space, maximum episode lengths and reward functions for different environments.}
\label{tab:env_reward_fns}
\end{table}

%% file: appendix/hyperparams.tex
\section{Experimental Details} \label{sec:hyperparameters}

Among the set of hyperparameters in our framework, the two crucial ones are the range of action duration, which is specified by $\dtmin$ and $\dtmax$, and the planning horizon. 
Rather than controlling both these values, $\dtmin$ is always set to $\dtenv$, and modifying $\dtmax$ controls the range.

\textbf{Details for Cartpole, Mountain Car and MuJoCo : }
For the environments Cartpole and Mountain Car, we use values of planning horizon from prior work for the standard framework, and adjust the depth for $A_{\tea}$ by exploring related values. 
We could not do this for MuJoCo-based environments since we set the frameskip parameter to 1. Rather we searched over some potential values for the planning horizon for the standard framework, and $\dtmax$ and planning horizon for our framework and chose the configuration with the best performance.
The other hyperparameters related to online training have been borrowed from \cite{chua2018deep}.

\input{tables/appendix/hyperparams}

\textbf{Details for Dubins Car environment : }
The default timescale for the environment is $0.2$ while $\dtmax$ for our formulation is 20. Setting $\dtmax$ to such a large value allows us to have the same value across maps and just allow more time for optimization. The planning horizon needs to be tuned for different maps. \Cref{tab:environment_parameters} contains the planning horizon for the maps used in the experiments. 
Note that for our experiment with the cave-mini map, we reduce $\dtmax$ to prevent the agent from completing the map in a few decision steps.

\input{images/appendix_env_images/main}

\input{appendix/model_learning}

\input{appendix/cem}

%% file: tables/appendix/hyperparams.tex
\begin{table}[tbph]
    \centering
    \scriptsize
    \begin{tabular}{lccccc}
    \toprule
        Domain &\multicolumn{2}{c}{Range of action duration  } & \multicolumn{2}{c}{Planning horizon } & Learning Rate \\
 \cmidrule{2-5}
    &  Standard & Ours  & Standard &  Ours    \\
    \midrule
Dubins Car [u-shaped map]       & 0.2 & [0.2-20] & 1000 & 75 & -\\
Dubins Car [cave-mini map]      & 0.2 & [0.2-2] & 120 & 50 & -\\
Cartpole                        & 1 & [1-10] & 30 & 3 & 1e-3\\
Mountain Car                    & 1 & [1-100] & 100 & 10 & -\\
IPC Mountain Car                & 1 & [1-125] & 175 & 12 & -\\
Reacher                         & 0.01 & [0.01, 0.2] & 25 & 5 & 1e-3\\
Pusher                          & 0.01 & [0.01, 0.2] & 25 & 5 & 1e-3\\
Half Cheetah                    & 0.01 & [0.01, 0.07] & 90 & 15 & 1e-3\\
Ant                             & 0.01 & [0.01, 0.07] & 70 & 15 & 3e-4\\
Hopper                          & 0.002 & [0.002, 0.125] & 50 & 15 & 3e-4\\
Walker                          & 0.002 & [0.002, 0.0625] & 70 & 15 & 3e-4\\
\bottomrule
\end{tabular}
\smallskip
\caption{Hyper-parameters for different environments. Range of action duration is for the fixed variant of the algorithm.}
\label{tab:environment_parameters}
\end{table}

%% file: images/appendix_env_images/main.tex
\begin{figure}[t]%
    \centering%
    \begin{subfigure}[b]{0.22\textwidth}
         \centering
           \includegraphics[width=\linewidth]{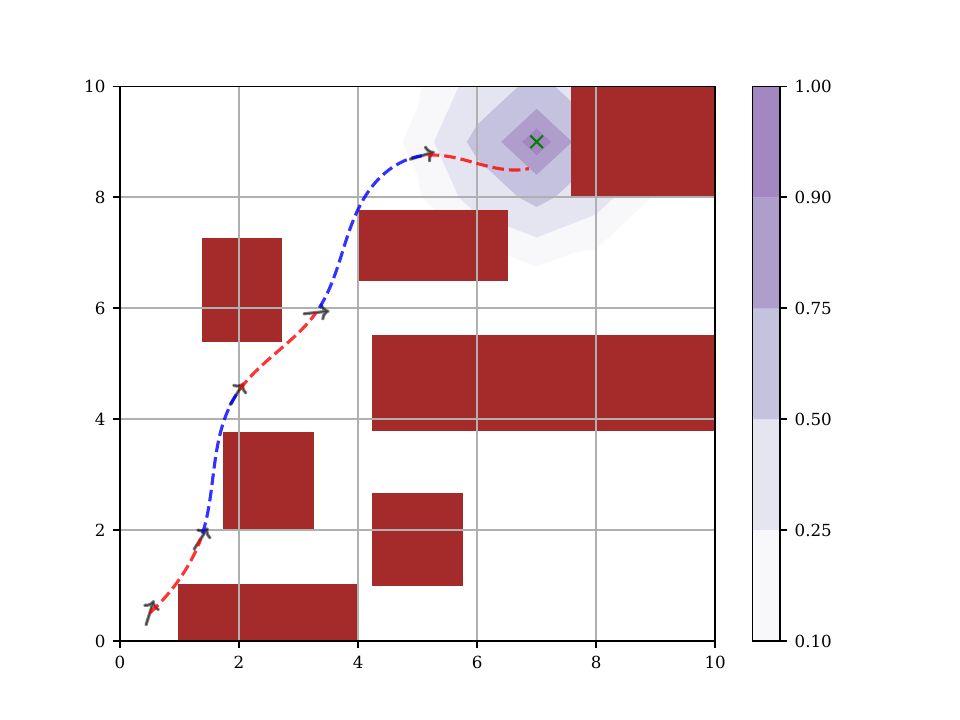}%
           \caption{}
           \label{fig:cave_mini_example}
    \end{subfigure}
    \begin{subfigure}[b]{0.22\textwidth}
         \centering
           \includegraphics[width=\linewidth]{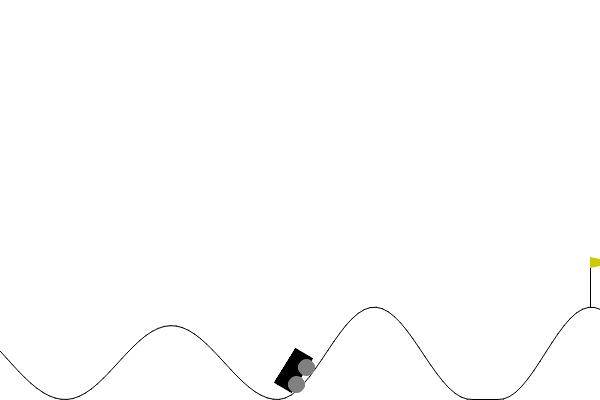}%
        \caption{}
        \label{fig:ipc_mountain_car_example}
    \end{subfigure}
    \caption{(a) An example of $A_{\tea}$ solving the cave-mini map $(\gamma_1 = 0.99, \gamma_2 = 1.0, \dtmax=20)$. (b) An instance of the IPC Multi-hill Mountain Car.}%
\end{figure}

%% file: appendix/model_learning.tex
\textbf{Model learning : }
For MBRL, we use the same model architecture as \cite{chua2018deep}. We learn an ensemble of 5 models, where each model is a fully connected neural network. For Ant, Half Cheetah and Hopper, the model has 4 hidden layers while for all other environments, it has 3 hidden layers. Each hidden layer has 200 neurons. The learning rates for each environment are given in \Cref{tab:environment_parameters}.

%% file: appendix/cem.tex
\input{algorithm/cem}

\textbf{Planning with an ensemble of models :}
We use the PETS framework from \citet{chua2018deep}. They introduce Trajectory Sampling (TS) which uses CEM as the planner. They maintain a sequence of sampling distributions from which $N$ action sequences are generated, and for every action sequence, they create $P$ particles.
Each particle is propagated using a particular member of the ensemble. For propagating particle $p$, TS1 uniformly resamples an ensemble member at each time step, while TS$\infty$ samples an ensemble member before an episode and uses that to propagate particle $p$ for the entire duration of the episode. The mean reward is computed for each action sequence. Then, the mean and the variance of the top $k$ action sequences gives the updated sampling distribution. Formally,
\begin{align} \label{eq:appendix_cem}
    \mu_{a} &= \frac{1}{k} \sum_{a \in \mathcal{K}} a \\
    \var_{a} &= \frac{1}{k} \sum_{a \in \mathcal{K}} (a - \mu_{a})^2
\end{align}
where $\mathcal{K}$ is the set of top $k$ action sequences.
This process is repeated for a fixed number of steps. 
For our experiments, we learn an ensemble of 5 feed-forward networks and use the TS$\infty$ variant with $P=20$ for planning. In \Cref{alg:planning_using_cem_and_ensemble_of_models}, we present the pseudocode for using TS$\infty$ with an ensemble of dynamics models.

%% file: algorithm/cem.tex
\begin{algorithm}[t]
\caption{Planning using TS$\infty$ with an ensemble of dynamics model.} \label{alg:planning_using_cem_and_ensemble_of_models}
\begin{algorithmic}[1]
\Require current state $(s)$, dynamics model ensemble$(\hat{f} = \{\hat{f}_1, \dots, \hat{f}_K\})$, planning horizon $(D)$, initial action distribution $(\mu_a, \var_a)$
\For{$i = 1$ to number of updates}
\State{Sample $N$ action sequences $\{a^{N}_1, a^{N}_{2}, \dots, a^{N}_{D}\}$} using $\mu_a, \sigma_a$.
 \For{$n = 1$ to $N$}
 \State Initialize $P$ particles = $\{a^{n,P}_1, a^{n,P}_{2}, \dots, a^{n,P}_{D}\}$.
 \For{$p = 1$ to $P$}
 \State \textcolor{blue}{\texttt{//propagate each particle using one model from ensemble}}
 \State{$j = p \texttt{ mod } K$} 
 \State{Propagate particle $p$ using $s^{n,p}_{t+1} = \hat{f}_j(s^{n,p}_t, a^{n,p}_t)$} and compute aggregate reward.
 \EndFor
 \State{Compute mean reward for action sequence using $\sum_{t=1}^{D} \sum_{p=1}^{P} \nicefrac{r(s^{n,p}_t, a^{n,p}_t)}{P}$}
  \EndFor
 \State{Choose the top $k$ action sequences and recompute $\mu_a, \var_a$}
\EndFor
\State{Execute first action from the optimal action sequence.}
\end{algorithmic}
\end{algorithm}

%% file: appendix/computational_resources.tex
\section{Computational Resources} \label{sec:computational_resources}

Each MBRL experiment was performed on a single node with a single GPU (using a mix of V100 and A100), single CPU (AMD EPYC 7742) and 64GB of RAM. The time required for training for different algorithms are in \Cref{tab:wall_times}. We note that although the experiments were done on two different types of GPUs, the running times for both of them are comparable.

\input{tables/appendix/wall_times.tex}

In Reacher and Pusher, the rollout length is mostly same for the standard algorithm and our proposed algorithm, but the planning horizon for the standard algorithm is 25 while for our proposed algorithm is 15, resulting in shorter training times. 

In Ant, Half Cheetah, Hopper and Walker $A_{\tea}$ is significantly faster than $A_{\naive}$. 
The dynamic variant of our proposed algorithm takes slightly longer than the fixed variant as the rollout length of the dynamic variant is longer on average than the rollout length of the fixed variant. 

%% file: tables/appendix/wall_times.tex
\begin{table*}[tbhp]
\centering
\scriptsize
\begin{tabular}{lccc}
\toprule
Env & Standard & Ours (F) & Ours (D) \\
\midrule
Cartpole & 0.65 & $\bm{0.4}$ & $\bm{0.5}$ \\
Half Cheetah & 45 & $\bm{5}$ & $\bm{5.5}$\\
Ant & 41 & $\bm{4}$ & $\bm{7}$ \\
Hopper & 40 & $\bm{4}$ & $\bm{6.5}$ \\
Reacher & 2.6 & $\bm{1.6}$ & $\bm{1.5}$ \\
Pusher & 2.6 & $\bm{2}$ & $\bm{1.5}$ \\
Walker & 36.8 & $\bm{3.1} $ & $\bm{5.8} $\\
\bottomrule
\end{tabular}
\caption{Approximate time (in hours) required to finish training for each environment.}
\label{tab:wall_times}
\end{table*}

%% file: appendix/additional_planning_experiments.tex
\section{Additional Planning Experiments}
\input{appendix/exp_ipc_mc}

\input{appendix/exp_u_maps}

%% file: appendix/exp_ipc_mc.tex
\subsection{Experiments with IPC Mountain Car} \label{sec:appendix_exp_with_ipc_mc}

\paragraph{How does the performance change if the increase the planning horizon?}
We vary the planning horizon of the agents and compare the performances across instances in \Cref{tab:exp_icaps_mc_varying_depth}. 
For the easier problems, a small planning horizon is sufficient, but for the difficult instances, a deeper search is required. The performance of the planner remains relatively stable as the planning horizon increases. 

\input{tables/exp_perf_vs_depth/icaps_mc_undiscounted_max_steps_300}

%% file: tables/exp_perf_vs_depth/icaps_mc_undiscounted_max_steps_300.tex
\begin{table*}[tbh]
\centering
\scriptsize
\begin{tabular}{cclllll}
\toprule
& D &  Instance 1 & Instance 2 & Instance 3 & Instance 4 & Instance 5\\
\midrule
\multirow{11}{*}{\raisebox{-\heavyrulewidth}{$A_{\naive}$}}
& 60    & 37.49 $\pm$ 51.36 & -0.0 $\pm$ 0.0 & -0.0 $\pm$ 0.0 & -0.0 $\pm$ 0.0 & -0.0 $\pm$ 0.0 \\
& 80    & 74.54 $\pm$ 41.7 & -0.0 $\pm$ 0.0 & -0.0 $\pm$ 0.0 & -0.0 $\pm$ 0.0 & -0.0 $\pm$ 0.0 \\
& 100   & 91.93 $\pm$ 0.41 & -0.0 $\pm$ 0.0 & -0.0 $\pm$ 0.0 & -0.0 $\pm$ 0.0 & -0.0 $\pm$ 0.0 \\
& 125   & 90.92 $\pm$ 0.22 & -0.0 $\pm$ 0.0 & -0.0 $\pm$ 0.0 & -0.0 $\pm$ 0.0 & -0.0 $\pm$ 0.0 \\
& 150   & 91.07 $\pm$ 0.18 & -0.0 $\pm$ 0.0 & -0.0 $\pm$ 0.0 & -0.0 $\pm$ 0.0 & -0.0 $\pm$ 0.0 \\
& 175   & 91.12 $\pm$ 0.3 & -0.0 $\pm$ 0.0 & -0.0 $\pm$ 0.0 & -0.0 $\pm$ 0.0 & -0.0 $\pm$ 0.0 \\
& 200   & 91.11 $\pm$ 0.27 & -0.0 $\pm$ 0.0 & -0.0 $\pm$ 0.0 & -0.0 $\pm$ 0.0 & -0.0 $\pm$ 0.0 \\
& 225   & 91.2 $\pm$ 0.32 & -0.0 $\pm$ 0.0 & -0.0 $\pm$ 0.0 & -0.0 $\pm$ 0.0 & -0.0 $\pm$ 0.0 \\
& 250   & 91.01 $\pm$ 0.12 & -0.0 $\pm$ 0.0 & -0.0 $\pm$ 0.0 & -0.0 $\pm$ 0.0 & -0.0 $\pm$ 0.0 \\
& 275   & 90.98 $\pm$ 0.11 & -0.0 $\pm$ 0.0 & -0.0 $\pm$ 0.0 & -0.0 $\pm$ 0.0 & -0.0 $\pm$ 0.0 \\
& 300   & 90.91 $\pm$ 0.1 & -0.0 $\pm$ 0.0 & -0.0 $\pm$ 0.0 & -0.0 $\pm$ 0.0 & -0.0 $\pm$ 0.0 \\
\midrule
\multirow{11}{*}{\raisebox{-\heavyrulewidth}{$A_{\tea}$}} 
& 2   & -0.01 $\pm$ 0.01 & -0.0 $\pm$ 0.0 & -0.0 $\pm$ 0.0 & -0.0 $\pm$ 0.0 & -0.0 $\pm$ 0.0 \\
& 5   & 73.52 $\pm$ 43.45 & -0.0 $\pm$ 0.0 & -0.0 $\pm$ 0.0 & -0.0 $\pm$ 0.0 & -0.0 $\pm$ 0.0 \\
& 8   & 92.58 $\pm$ 1.01 & 90.99 $\pm$ 0.7 & -0.0 $\pm$ 0.0 & -0.0 $\pm$ 0.0 & -0.0 $\pm$ 0.0 \\
& 10  & 71.43 $\pm$ 46.05 & 89.79 $\pm$ 1.28 & 87.27 $\pm$ 1.86 & 68.11 $\pm$ 42.49 & 85.02 $\pm$ 0.26 \\
& 12  & 92.41 $\pm$ 1.82 & 90.11 $\pm$ 1.19 & 87.44 $\pm$ 1.52 & 86.91 $\pm$ 1.53 & 85.45 $\pm$ 0.25 \\
& 15  & 92.33 $\pm$ 2.0 & 90.52 $\pm$ 0.6 & 87.91 $\pm$ 0.8 & 87.51 $\pm$ 0.98 & 85.19 $\pm$ 1.15 \\
& 20  & 92.52 $\pm$ 1.46 & 90.12 $\pm$ 1.36 & 87.74 $\pm$ 1.99 & 87.08 $\pm$ 2.57 & 86.43 $\pm$ 0.31 \\
& 25  & 92.39 $\pm$ 1.85 & 90.54 $\pm$ 1.06 & 88.39 $\pm$ 1.36 & 87.71 $\pm$ 1.61 & 87.06 $\pm$ 0.46 \\
& 30  & 72.07 $\pm$ 45.41 & 91.08 $\pm$ 0.37 & 88.94 $\pm$ 0.49 & 68.0 $\pm$ 45.24 & 86.08 $\pm$ 1.81 \\
& 35  & 51.48 $\pm$ 57.79 & 90.91 $\pm$ 1.15 & 89.28 $\pm$ 0.2 & 89.06 $\pm$ 0.81 & 67.07 $\pm$ 45.5 \\
& 40  & 71.21 $\pm$ 48.64 & 90.12 $\pm$ 1.42 & 89.75 $\pm$ 2.19 & 87.38 $\pm$ 2.38 & 67.86 $\pm$ 45.2 \\
\bottomrule
\end{tabular}
\caption{Performance of $A_{\naive}$ and $A_{\tea}$ in various instances of IPC Multi-hill Mountain Car as the planning horizon is increased.}
\label{tab:exp_icaps_mc_varying_depth}
\end{table*}

%% file: appendix/exp_u_maps.tex
\subsection{Experiments with U-shaped Maps in Dubins Car} \label{sec:appendix_exp_u_maps}

We experiment with varying depths using u-shaped maps in the Dubins Car environment. The problem is not trivial as the car faces the obstacle and it needs to first turn around and then find a path the goal, which gives the agent a reward of 100. Every collision with an obstacle gets a penalty of 10. An episode terminates when the agent reaches the goal or when the agent takes 300 primitive steps. 

To be successful in this setting, we observe that $A_{\naive}$ requires a planning horizon of 1000 and $A_{\tea}$ requires a planning horizon of 75. Both the agents use 10,000 samples and 50 optimization steps for each decision. 
$A_{\tea}$ succeeds but not in all runs. 
We look at the failure cases for $A_{\tea}$ and have an interesting observation. For planning horizon of 75 and 100, the failure is due to the fact that the agent runs out of primitive actions, while for planning horizon 50, the failing scenario corresponds to the agent not being able to find a path out of the obstacles region. This means that if we allow the episodes to run for longer, the former failure can be mitigated but the latter cannot. The detailed results are in \Cref{tab:exp_u_maps_varying_conf} and the failure cases for $A_{\tea}$ are shown in \Cref{fig:exp_u_maps_failure_cases}.

\input{tables/exp_dubins_car_u_maps/u4_varying_conf}

\input{images/exp_u_maps_failure_cases/main}

Next, we make the problem difficult by augmenting the action space with 100 dummy action variables. As the agents are unaware that these variables don't contribute to the reward or the dynamics, they still need to search over these variables. We set up the experiment similar to the previous one by keeping the number of samples to 10000 and varying the planning horizon. For $A_{\naive}$, almost all the runs fail - either because it does not find a path around the obstacle, or crashes due to huge memory requirements. In order to reduce the memory requirements, we perform a second experiment, fixing the planning horizon to 1000 and varying the number of samples. Reducing the number of samples fixes the issue of memory requirements, but does not help the agent identify a path. 
In contrast, the performance of $A_{\tea}$ is mostly unaffected by this addition of dummy action variables. The detailed results are in \Cref{tab:exp_u_maps_varying_conf_hard}. $A_{\tea}$ uses the same configuration as the earlier experiment and achieves similar performance. As observed earlier, the failure case of $A_{\tea}$ for planning horizon of 50 is due to the fact that it cannot find a path around the obstacles, while the failure case for planning horizon of 100 is because it runs out of primitive steps.

%% file: tables/exp_dubins_car_u_maps/u4_varying_conf.tex
\begin{table*}[tbhp]
\centering
\scriptsize
\begin{tabular}{ccllllc}
\toprule
& D &  Rewards $\uparrow$ & Decision Steps $\downarrow$ & Decision time $\downarrow$ & Time for an episode $\downarrow$ & P(success) $\uparrow$\\
\midrule
\multirow{4}{*}{\raisebox{-\heavyrulewidth}{$A_{\naive}$}}
& 250     &  -9.95 $\pm$ 0.0    &  300.0 $\pm$ 0.0     & 0.13 $\pm$ 0.0     &  413.46 $\pm$ 4.16  & 0 \\
& 500     &  -9.95 $\pm$ 0.0    &  300.0 $\pm$ 0.0     & 0.29 $\pm$ 0.0     &  806.97 $\pm$ 8.82  & 0 \\
& 750     & 32.04 $\pm$ 62.17   &  213.8 $\pm$ 118.03  & 0.42 $\pm$ 0.02    & 851.53 $\pm$ 466.61 & 0.4 \\
& 1000    & 100.0 $\pm$ 0.0     &  90.8  $\pm$ 20.17   & 0.56 $\pm$ 0.02    & 480.38 $\pm$ 102.06 & 1 \\
\midrule

\multirow{4}{*}{\raisebox{-\heavyrulewidth}{$A_{\tea}$}}
& 25     &  -497.53 $\pm$ 137.16      &  4.0 $\pm$ 0.0   &  2.31 $\pm$ 0.09       &  14.1 $\pm$ 0.43    & 0 \\
& 50     &  16.32 $\pm$ 187.12        &  4.2 $\pm$ 0.45  &  3.66 $\pm$ 0.1        &  20.03 $\pm$ 1.59   & 0.8 \\
& 75     &  80.0 $\pm$ 44.72          & 4.4 $\pm$ 1.14  &   5.08 $\pm$ 0.24       &  27.39 $\pm$ 5.84   & 0.8 \\
& 100    &  80.0 $\pm$ 44.72          &  4.4 $\pm$ 1.14 &   6.57 $\pm$ 0.21       & 34.56 $\pm$ 7.79    & 0.8 \\
\bottomrule
\end{tabular}
\caption{Performance of $A_{\naive}$ and $A_{\tea}$ on the u-shaped map in Dubins Car as the planning horizon is varied. Both agents use 10,000 samples and 50 optimization steps.}
\label{tab:exp_u_maps_varying_conf}
\end{table*}

\begin{table*}[tbhp]
\centering
\scriptsize
\begin{tabular}{cccllllc}
\toprule
& D & \#samples &  Rewards $\uparrow$ & Decision Steps $\downarrow$ & Decision time $\downarrow$ & Time for an episode $\downarrow$ & P(success) $\uparrow$\\
\midrule
\multirow{4}{*}{\raisebox{-\heavyrulewidth}{$A_{\naive}$}}
& 250  & \multirow{4}{*}{\raisebox{-\heavyrulewidth}{10000}}   &  -9.95 $\pm$ 0.0    &  300.0 $\pm$ 0.0     &  0.66 $\pm$ 0.0     &  580.85 $\pm$ 2.91  & 0 \\
& 500  &    &  12.04 $\pm$ 49.17    &  255.2 $\pm$ 100.18     &  1.38 $\pm$ 0.01     &  970.11 $\pm$ 378.22  & 0.2 \\
& 750  &    & -   & -  & -    & - & 0 \\
& 1000 &   & -     &  -   & -    & - & 0 \\
\midrule
\multirow{3}{*}{\raisebox{-\heavyrulewidth}{$A_{\naive}$}}
& \multirow{3}{*}{\raisebox{-\heavyrulewidth}{1000}}  &  100   &  -919.43 $\pm$ 238.36    &  300.0 $\pm$ 0.0     & 0.46 $\pm$ 0.03     &  1567.29 $\pm$ 25.31  & 0 \\
&   & 1000   &   -15.92 $\pm$ 8.9    &   300.0 $\pm$ 0.0    &   0.73 $\pm$ 0.0      &   1663.75 $\pm$ 8.56  & 0 \\
&   & 10000   & -   & -  & -    & - & 0 \\
\midrule

\multirow{4}{*}{\raisebox{-\heavyrulewidth}{$A_{\tea}$}}
& 25    & \multirow{4}{*}{\raisebox{-\heavyrulewidth}{10000}} & -479.62 $\pm$ 145.46      &  4.0 $\pm$ 0.0   &  2.31 $\pm$ 0.09       & 14.05 $\pm$ 0.25    & 0 \\
& 50    & &  18.31 $\pm$ 182.67       &  3.6 $\pm$ 0.55    &  4.0 $\pm$ 0.12        &  18.72 $\pm$ 2.41   & 0.8 \\
& 75    & &  100.0 $\pm$ 0.0          &  4.6 $\pm$ 0.89    &  5.36 $\pm$ 0.16       &  30.01 $\pm$ 4.56   & 1.0 \\
& 100   & &   80.0 $\pm$ 44.72        &  4.4 $\pm$ 0.89    &  6.87 $\pm$ 0.12       &  35.74 $\pm$ 6.44    & 0.8 \\
\bottomrule
\end{tabular}
\caption{Performance of $A_{\naive}$ and $A_{\tea}$ on the u-shaped map in Dubins Car when the action space is augmented with 100 dummy actions. Both agents use 50 optimization steps. Missing values indicate that the particular configuration was not feasible due to large memory requirements.}
\label{tab:exp_u_maps_varying_conf_hard}
\end{table*}

%% file: images/exp_u_maps_failure_cases/main.tex
\begin{figure*}[tbhp]%
    \centering%
    \begin{subfigure}[b]{0.32\textwidth}
         \centering
           \includegraphics[width=\linewidth]{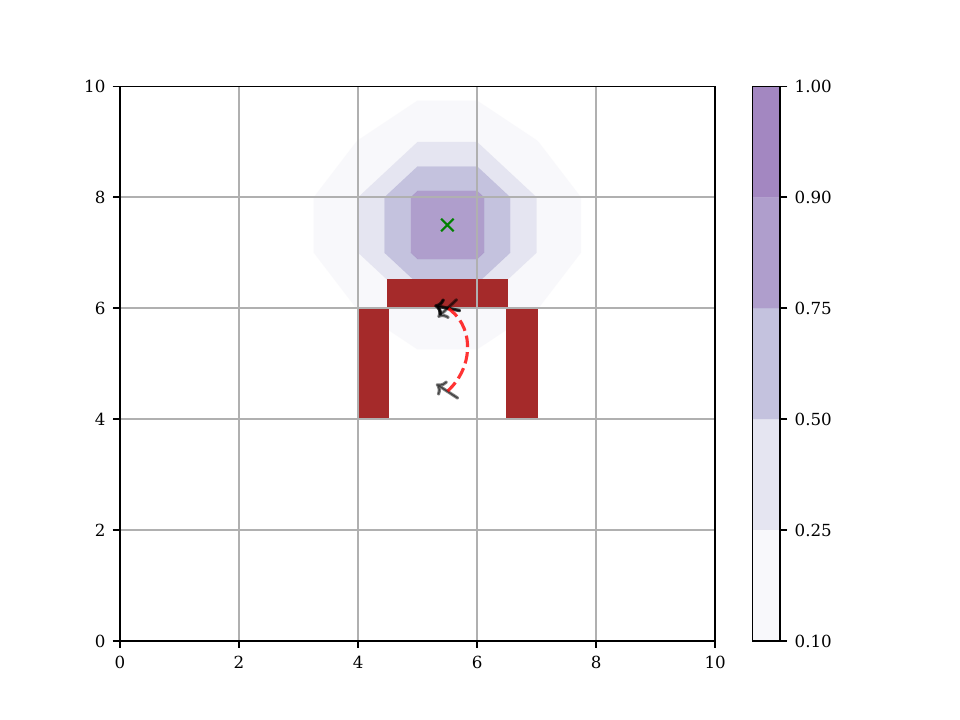}%
     \caption{$D_{\tea} = 50, |\A|=2$}
    \end{subfigure}
    \hfill
    \begin{subfigure}[b]{0.32\textwidth}
         \centering
           \includegraphics[width=\linewidth]{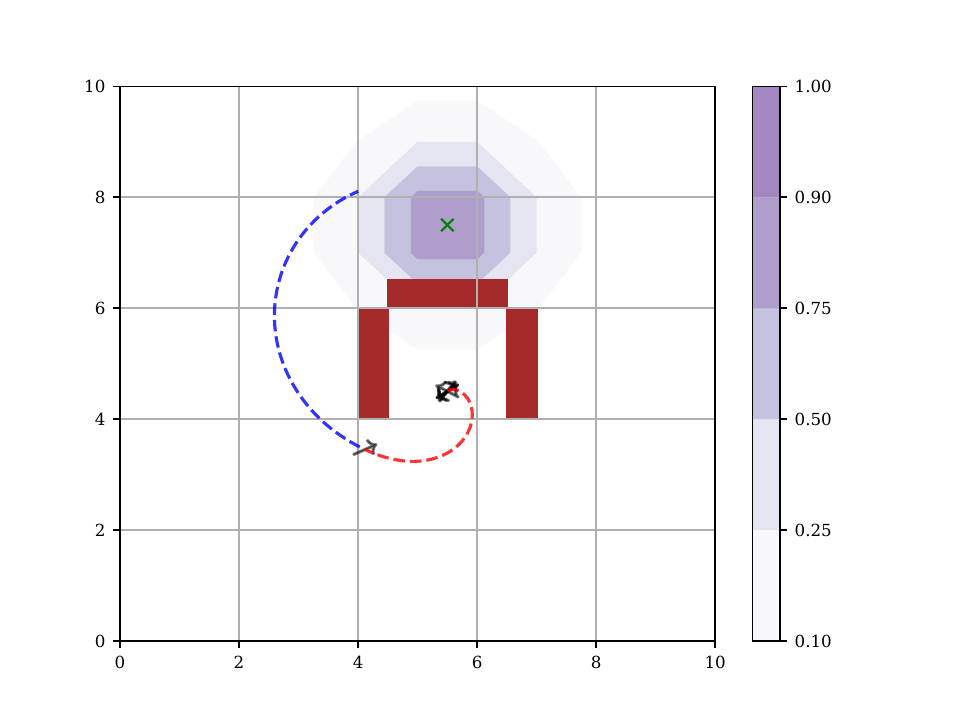}%
     \caption{$D_{\tea} = 75, |\A|=2$}
    \end{subfigure}
    \hfill
    \begin{subfigure}[b]{0.32\textwidth}
         \centering
         \includegraphics[width=\linewidth]{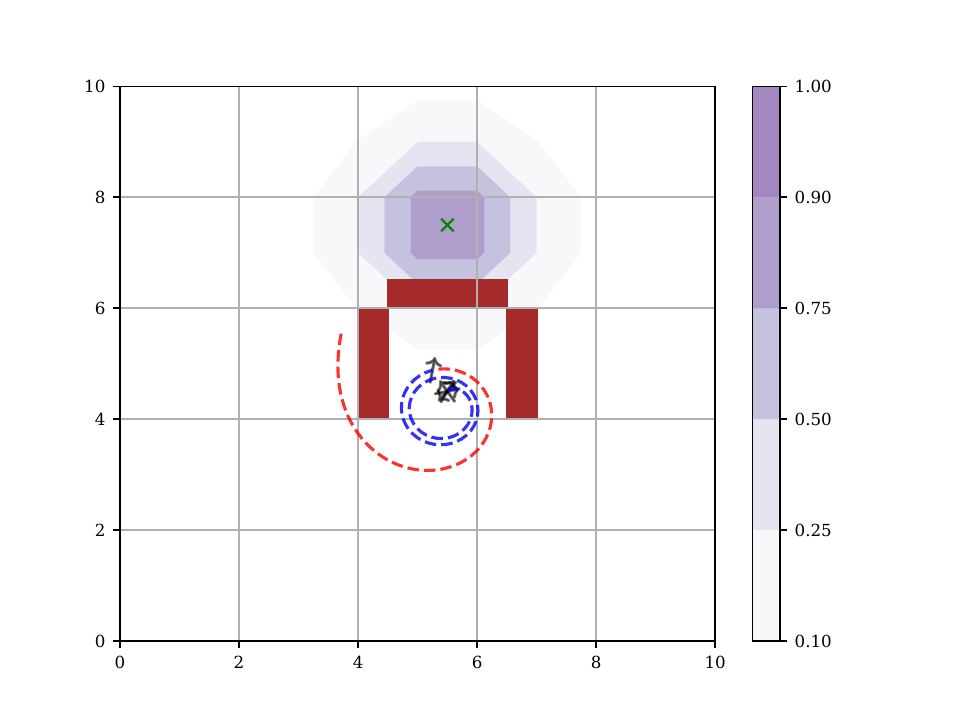}%
        \caption{$D_{\tea} = 100, |\A|=2$}
    \end{subfigure}
    \\
    \begin{subfigure}[b]{0.32\textwidth}
         \centering
           \includegraphics[width=\linewidth]{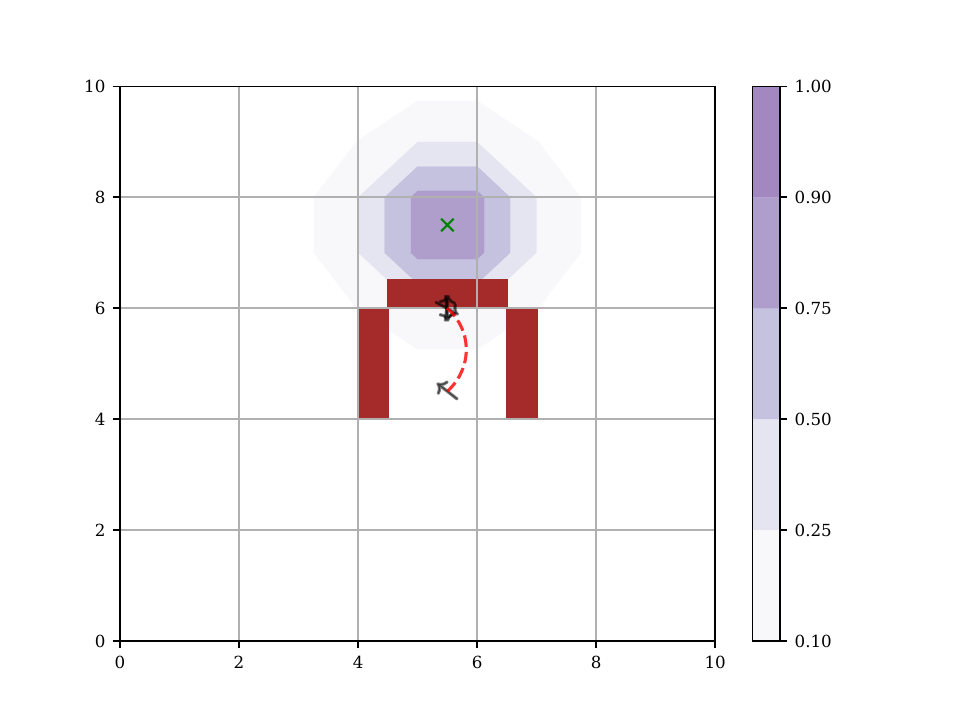}%
     \caption{$D_{\tea} = 50, |\A|=102$}
    \end{subfigure}
    \hfill
    \begin{subfigure}[b]{0.32\textwidth}
         \centering
           \includegraphics[width=\linewidth]{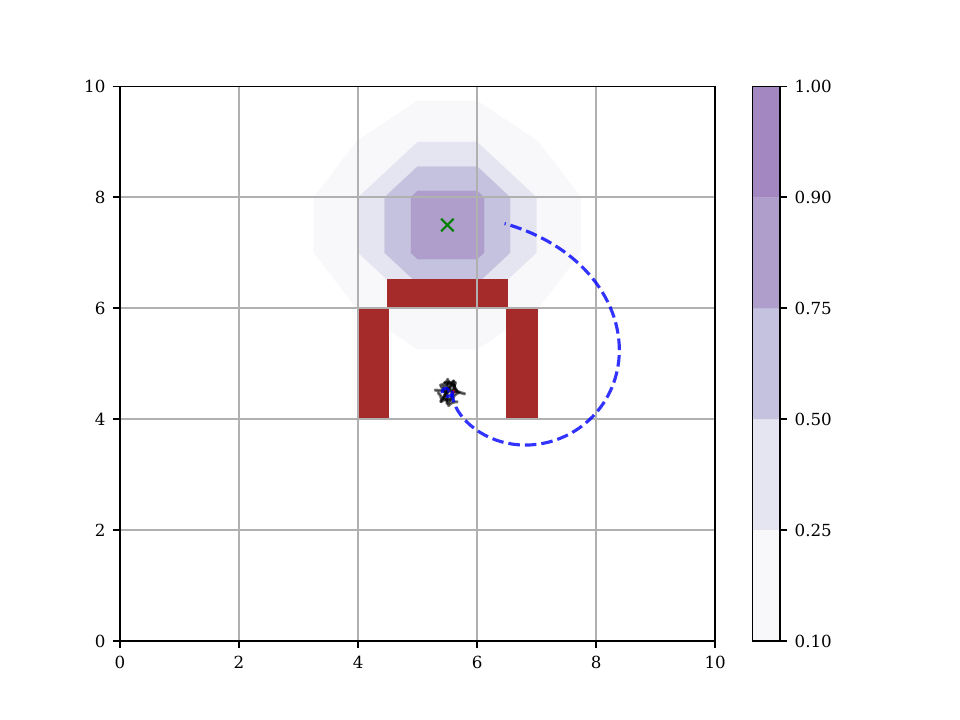}%
     \caption{$D_{\tea} = 100, |\A|=102$}
    \end{subfigure}
    \caption{Failure cases for $A_{\tea}$ in the u-shaped map. When $|\A|=2$, for $D_{\tea} = 75$ and $D_{\tea} = 100$, the agent identifies a path around the obstacles but runs out of the maximum number of primitive steps. On increasing the action space, a similar failure occurs for $D_{\tea} = 100$. The failures due to $D_{\tea}=50$ happen because the agent does not find a path around the obstacles.}
    \label{fig:exp_u_maps_failure_cases}
\end{figure*}

%% file: appendix/additional_mbrl_results.tex
\section{Additional MBRL results}

\subsection{How do the action repeats due to $A_{\tea}$ vary across training iterations?}
\input{images/exp_online_learning/appendix}

While the histogram in \Cref{fig:histogram_of_action_repeats} gives us a good sense of the distribution of discretized action duration $(\nicefrac{\dt}{\dtenv})$ chosen by $A_{\tea}$ in one episode after the training has been completed, it does not capture the trend across time. 
As the average action repeat for $A_{\tea}(\text{D})$ can have a large range, in \Cref{fig:learning_curves_online_learning_appendix} , we plot the log of average action repeat instead. 
$A_{\naive}$ has an action repeat of 1 as it works with primitive actions. For $A_{\tea}(\text{F})$, the average action repeat is mostly stable. In Cartpole, Reacher and Pusher, the agent has larger action repeats (or takes large values of $\dt$) in the beginning before realizing that smaller action repeats work better. The more interesting plots are those of $A_{\tea}(\text{D})$ as it has to identify the correct range of $\dt$. From the plots, we see that the average action repeats decreases gradually before stabilizing around the average action repeat used by $A_{\tea}(\text{F})$. Note that larger values of average action repeats does not necessarily mean smaller number of decision points. $A_{\tea}(\text{D})$ has to choose $\dtmax$ candidates from a list of exponentially spaced candidates and a large $\dtmax$ selection can easily skew the average action repeat.

%% file: images/exp_online_learning/appendix.tex
\begin{figure*}[tph]%
    \centering%
    \begin{subfigure}[b]{0.245\textwidth}
         \centering
           \includegraphics[width=\linewidth]{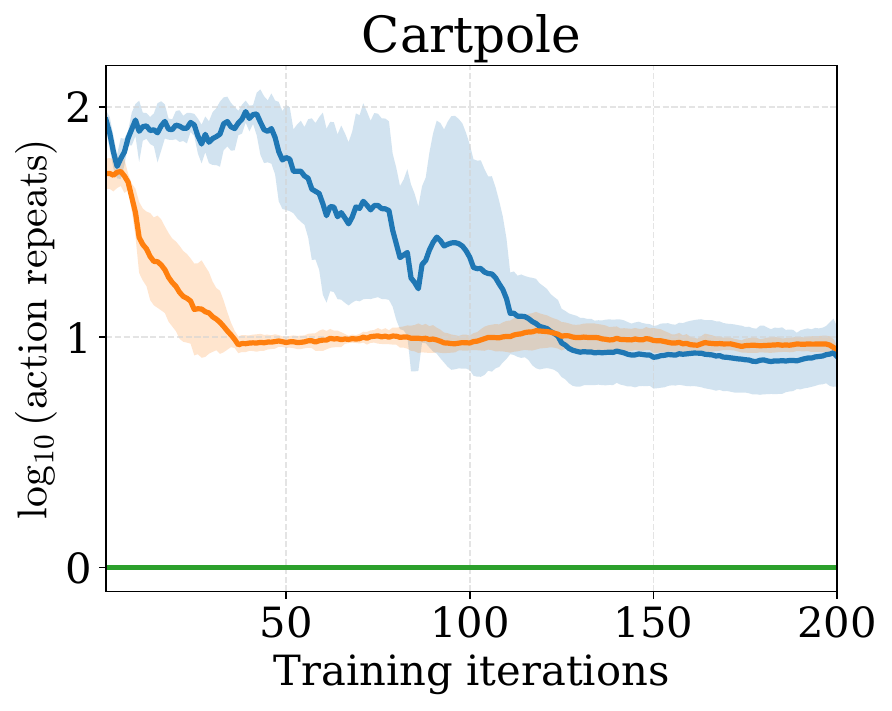}%
    \end{subfigure}
    \hfill
    \begin{subfigure}[b]{0.245\textwidth}
         \centering
           \includegraphics[width=\linewidth]{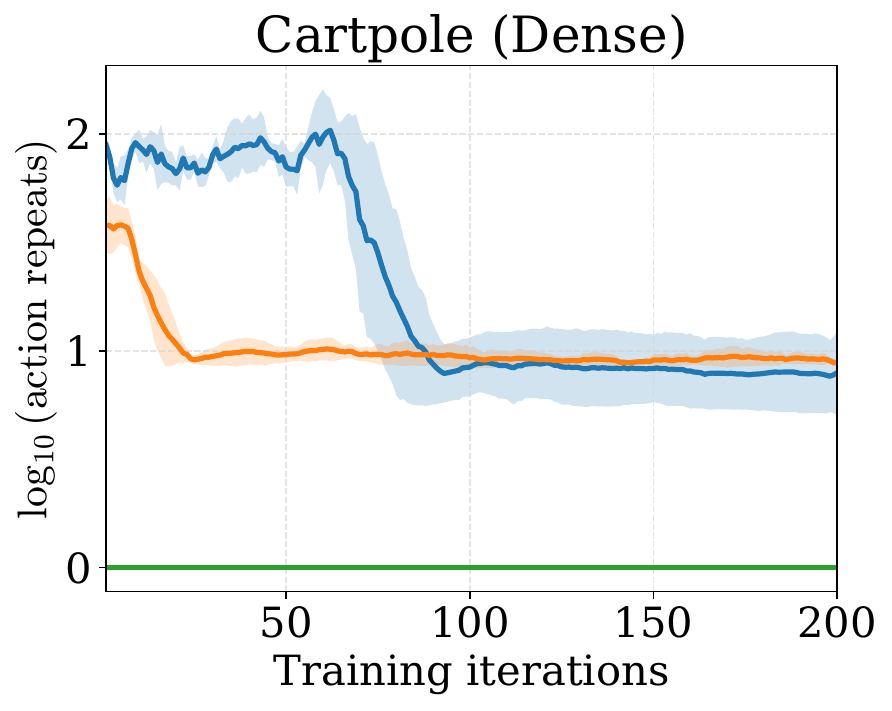}%
    \end{subfigure}
    \hfill
    \begin{subfigure}[b]{0.245\textwidth}
         \centering
           \includegraphics[width=\linewidth]{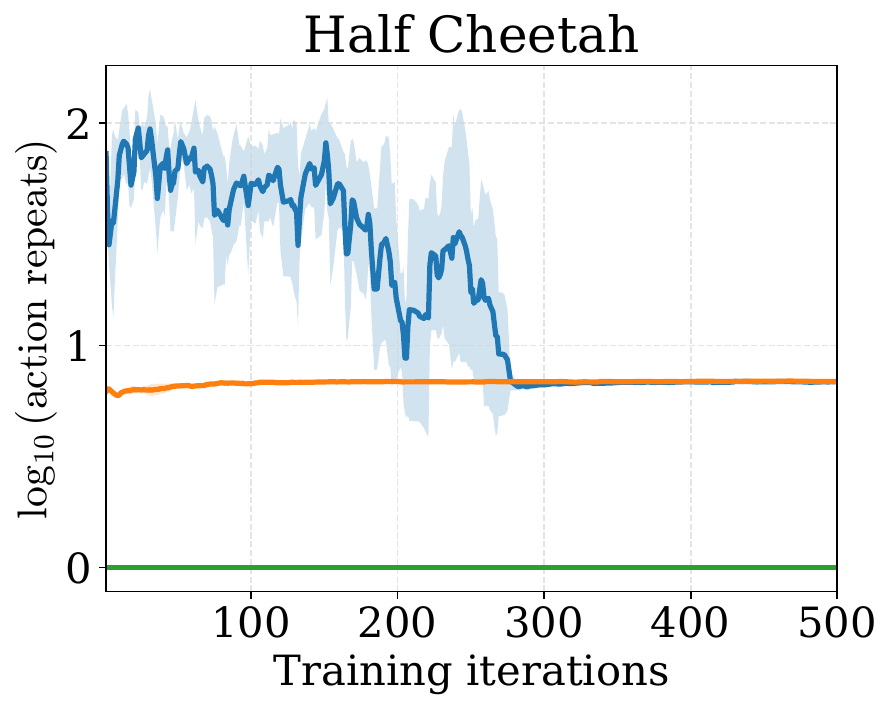}%
    \end{subfigure}
    \hfill
    \begin{subfigure}[b]{0.245\textwidth}
         \centering
           \includegraphics[width=\linewidth]
           {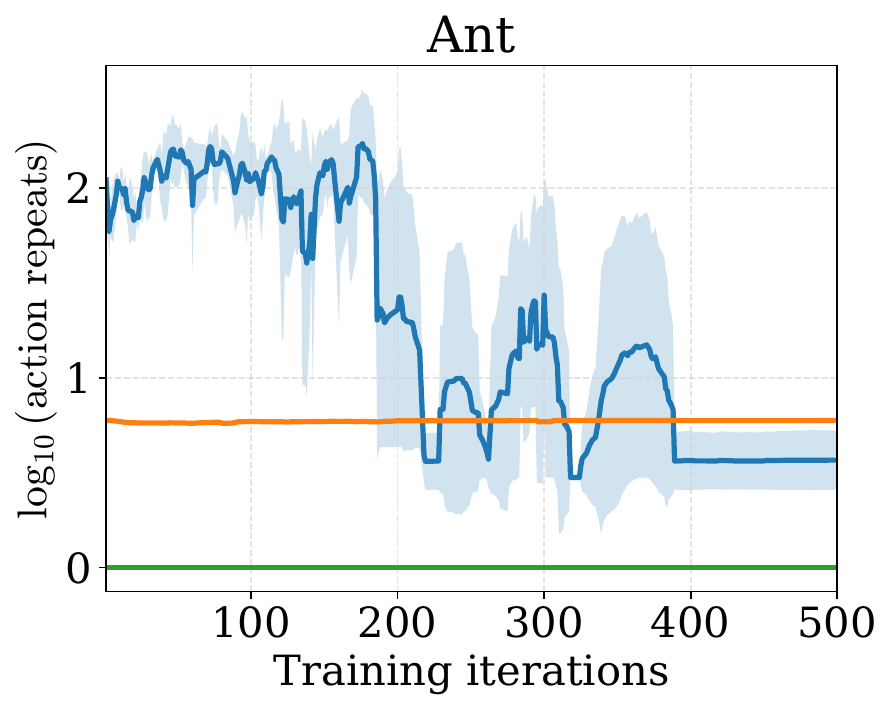}%
    \end{subfigure}
    \\
    \begin{subfigure}[b]{0.245\textwidth}
         \centering
           \includegraphics[width=\linewidth]
           {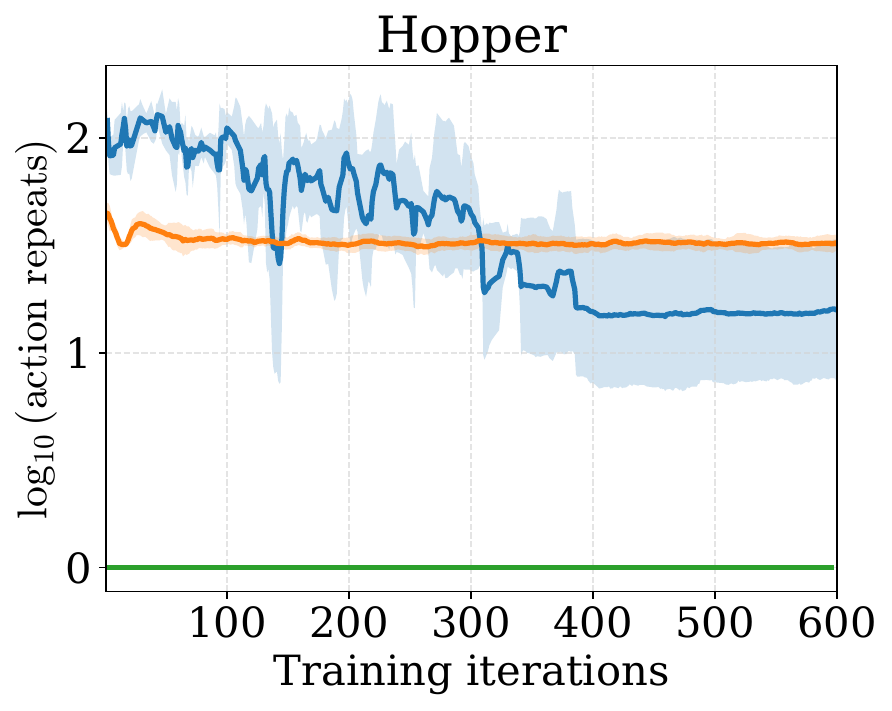}%
    \end{subfigure}
      \begin{subfigure}[b]{0.245\textwidth}
         \centering
           \includegraphics[width=\linewidth]
           {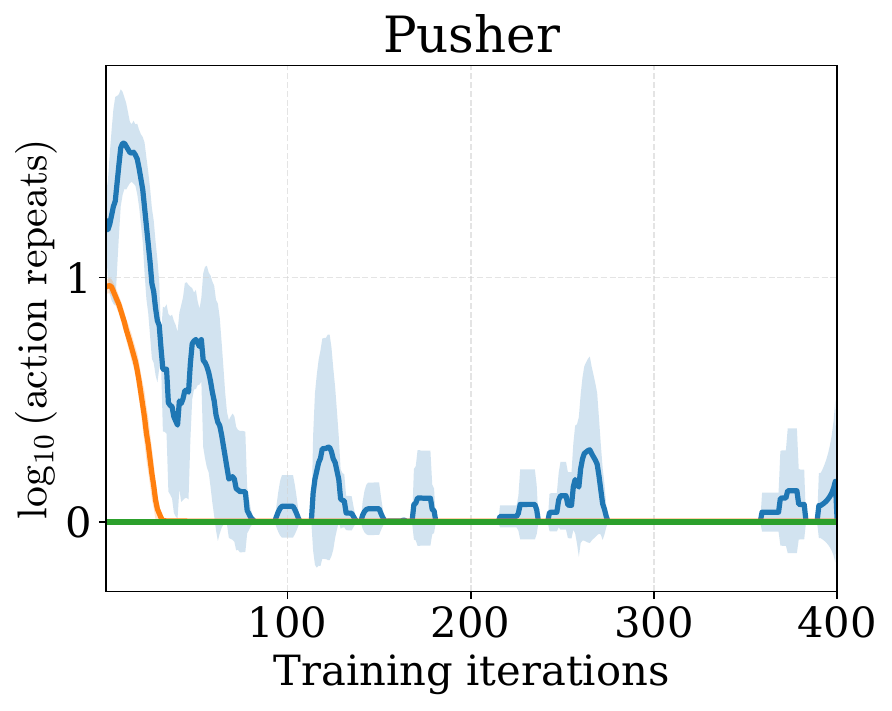}%
    \end{subfigure}
          \begin{subfigure}[b]{0.245\textwidth}
         \centering
           \includegraphics[width=\linewidth]
           {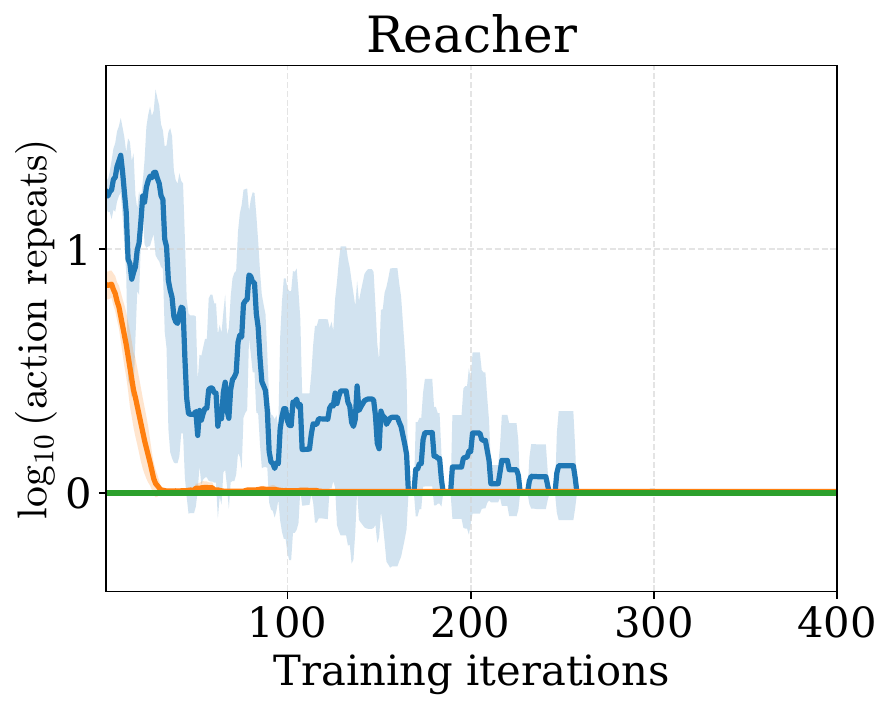}%
    \end{subfigure}
          \begin{subfigure}[b]{0.245\textwidth}
         \centering
           \includegraphics[width=\linewidth]
           {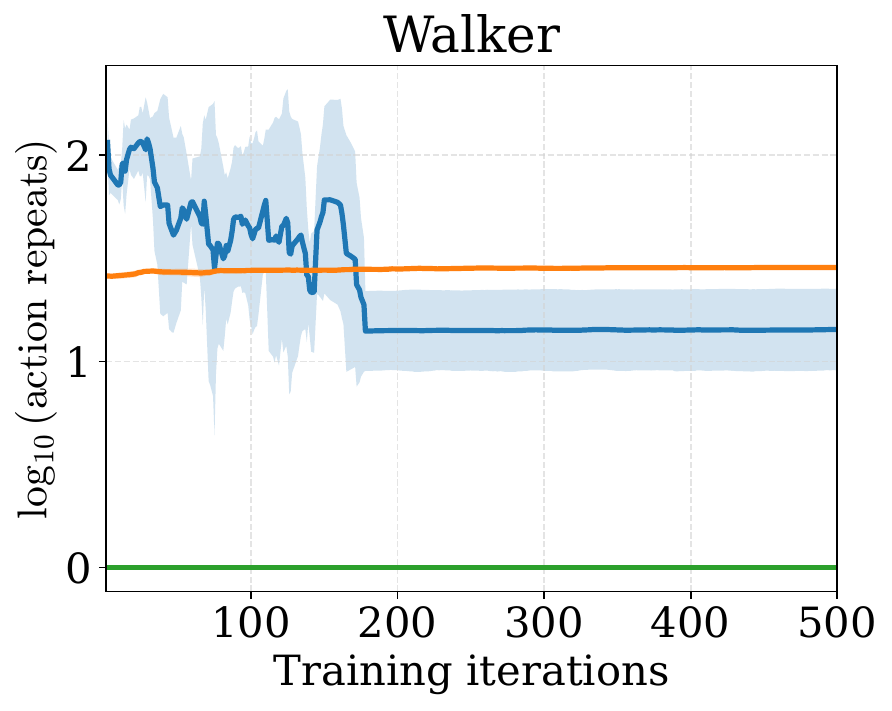}%
    \end{subfigure}
    \\
    \begin{subfigure}[b]{0.33\textwidth}
         \centering
         \includegraphics[width=\linewidth]{images/exp_online_learning/legend.pdf}%
    \end{subfigure}
    
    \caption{Mean and standard deviation of the running averages (window size=10) of log of action repeats. Values due to $A_{\tea}(\text{D})$ converge in the neighborhood of values due to $A_{\tea}(\text{F})$}.
    \label{fig:learning_curves_online_learning_appendix}
\end{figure*}

%% file: appendix/ablation/main.tex
\section{Ablations}
\input{appendix/ablation/exp_single_model_vs_sep_models}
\input{appendix/ablation/comparison_model_prediction.tex}

%% file: appendix/ablation/exp_single_model_vs_sep_models.tex
\subsection{Single model vs. Separate models}
\label{sec:appendix_single_model_vs_sep_model}
In non-stationary bandits, the reward distribution of the arms changes with time. If the reward distribution of an arm changes only when it is pulled, it is called a \emph{rested} bandit, while if the reward distribution of an arm changes irrespective of whether that particular arm was pulled or not, it is called a \emph{restless} bandit \citep{whittle1988restless,tekin2012online}.

\input{images/exp_single_model_vs_sep_model/main}

In the MBRL setup, the performance of the agent depends on the quality of the learned dynamics model. If we learn a single dynamics model to be shared across all arms, then the underlying reward distribution for an arm can keep on changing even if the arm is never pulled. This is the setting of a restless bandit. On the other hand, learning a separate dynamics model for each arm results in a rested bandit setting as it ensures that the underlying reward distribution for an arm only changes when that arm is pulled and the corresponding dynamics gets updated. 

Hence, choosing whether to use a single dynamics model which is shared between all the bandit arms or using a separate dynamics model for each bandit is an important consideration. Both the modeling choices have their advantages and disadvantages.

While using a single dynamics makes the most use of the available data, there is a data imbalance issue. 
Larger values of $\dt$ will result in fewer experiences being captured. Even if we uniformly sample the arms, we will have fewer examples due to candidates with large values of $\dtmax$ than we will have for candidates with small values of $\dtmax$. As all of the data is used to train a single model, the model is likely to overfit on examples from candidates with small values of $\dtmax$.
On the other hand, using separate dynamics model for each bandit makes the model more focused, but we do not make full use of the available data. 

As shown in \Cref{fig:learning_curves_single_model_vs_sep_model}, we find that using the UCB heuristic, the agent performs better if it learns a separate dynamics model for each bandit arm.

%% file: images/exp_single_model_vs_sep_model/main.tex
\begin{figure*}[b]%
    \centering%
    \begin{subfigure}[b]{0.3\textwidth}
         \centering
           \includegraphics[width=\linewidth]{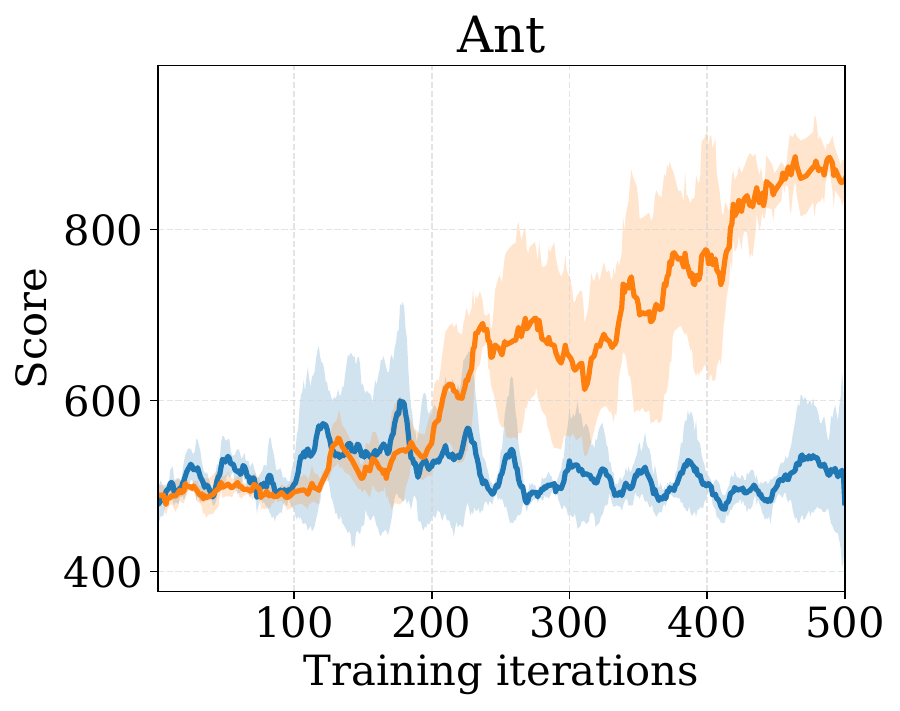}%
    \end{subfigure}
    \begin{subfigure}[b]{0.3\textwidth}
         \centering
           \includegraphics[width=\linewidth]{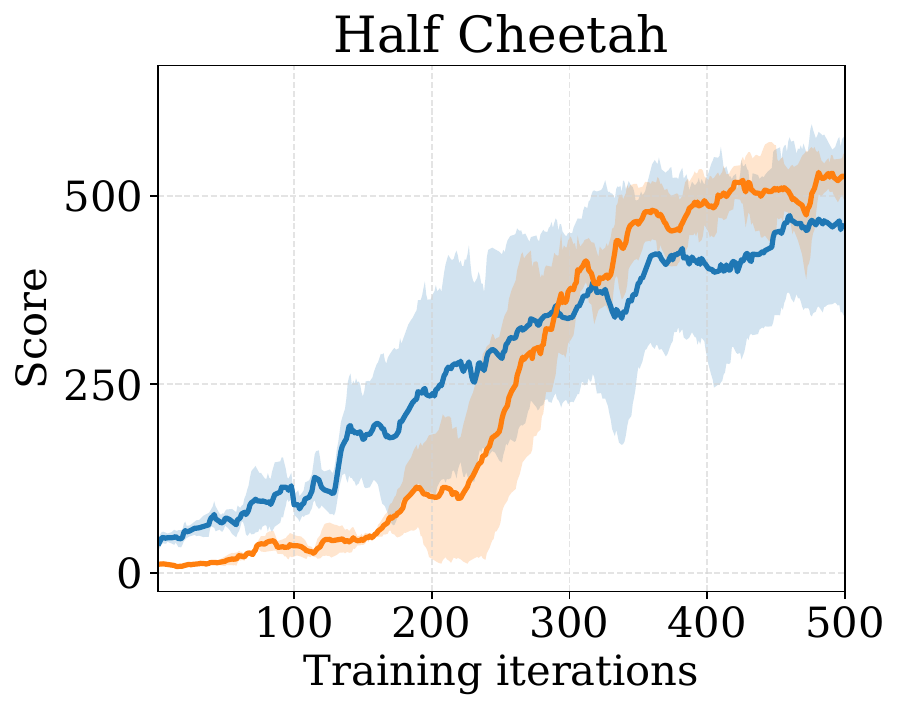}
    \end{subfigure}
    \hfill
    \\
    \begin{subfigure}[b]{0.3\textwidth}
         \centering
           \includegraphics[width=\linewidth]{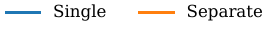}%
    \end{subfigure}
    \caption{While learning a single dynamics model to be shared across all $\dtmax$ candidates is sufficient for Half Cheetah, we need to learn separate dynamics models for each $\dtmax$ candidate in Ant.}
    \label{fig:learning_curves_single_model_vs_sep_model}
\end{figure*}

%% file: appendix/ablation/comparison_model_prediction.tex
\subsection{Which is a better approximation for $F$ - $\fhattea$ or $\fhatip$?}
While we proposed to use $\fhattea$ to approximate the temporally-extended dynamics function $F$, one could also consider using $\fhatip$ which approximates the one-step dynamics function $f$ and then iterates over it. While we know that $\fhattea$ leads to better time complexity than $\fhatip$, it is interesting to explore how it compares in terms of prediction quality.

To answer this question, we perform two experiments with random action sequences and compare the prediction quality of $\fhattea$ and $\fhatip$. In both the experiments, we sample random action sequences of length 500 and compute the mean-squared error~(MSE) between the actual states observed upon taking the action sequences in the environment and the states as predicted by the learned model.  The difference between the two experiments lies in the type of action sequences being sampled. For the first experiment, we randomly sample one-step action sequences which implicitly correspond to duration $\dtmin$, while for the second experiment, we randomly sample temporally-extended action sequences with action duration sampled from $\mathcal{U}(\dtmin, \dtmax)$. Each experiment was repeated 10 times in order to get a better estimate and the mean results are shown in \Cref{tab:comparison_model_prediction}.

\input{tables/appendix/comparison_model_prediction.tex}

In Ant, when predicting using temporally-extended action sequences, in 4 out of 10 experiments, the MSE due to $\fhatip$ was extremely high. We hypothesize that the states observed during these experiments were probably outside the range of examples seen during the training of the one-step model that is used by $\fhatip$. $\fhatip$ without outliers reports the MSE after removing these 4 experiments.

As can be observed, for one-step sequences $\fhatip$ has a lower MSE but for temporally-extended sequences, $\fhattea$ is clearly the better choice. This is not surprising since using temporally-extended action sequences with $\fhatip$ requires iterating over the one-step model multiple times which can lead to compounding errors. Thus, $\fhattea$ is a better choice both in terms of time complexity and prediction quality when using temporally-extended action sequences.

%% file: tables/appendix/comparison_model_prediction.tex
\begin{table*}[h]
\centering
\scriptsize
\begin{tabular}{l|cc|cc|ccc}
\toprule
& & &\multicolumn{2}{c|}{Random one step sequences} & \multicolumn{3}{c}{Random temporally extended sequences} \\
Env & $\dtmin$ & $\dtmax$ & $\fhattea$ & $\fhatip$ & $\fhattea$ & $\fhatip$ & $\fhatip$ without outliers \\
\midrule
Cartpole & 1 & 10 & 0.0107 & $2.517\times10^{-7}$ & 0.0184 & 0.065 & 0.065 \\
Ant & 0.01 & 0.07 & 0.126 & 0.0149 & 0.298 & $4.525 \times 10^{10}$ & 0.60 \\
Half Cheetah & 0.01 & 0.07 & 0.30 & 0.04 & 1.752 & 2.526 & 2.526 \\
\bottomrule
\end{tabular}
\caption{Comparison between the prediction error (in terms of MSE) of $\fhattea$ and $\fhatip$ on random action sequences.}
\label{tab:comparison_model_prediction}
\end{table*}